\documentclass[11pt, a4paper, copyright, goog]{google}

\usepackage[authoryear, sort&compress, round]{natbib}
\usepackage[utf8]{inputenc} % allow utf-8 input
\usepackage[T1]{fontenc}    % use 8-bit T1 fonts
\usepackage{hyperref}       % hyperlinks
\usepackage{url}            % simple URL typesetting
\usepackage{booktabs}       % professional-quality tables
\usepackage{amsfonts}       % blackboard math symbols
\usepackage{nicefrac}       % compact symbols for 1/2, etc.
\usepackage{microtype}      % microtypography
\usepackage{graphicx}
\usepackage[ruled,vlined]{algorithm2e}
\usepackage{xcolor}
\usepackage{caption}
\usepackage{subcaption}
\usepackage{amsmath}
\usepackage{authblk}
\usepackage{doi}


\uselogo{} 

\title{Coevolution of self-replication and function in a digital primordial soup}

\correspondingauthor{fcicala@google.com, blakerichards@google.com}

\author[*,1]{Francesco Cicala}
\author[*,1]{Eyvind Niklasson}
\author[*,1]{Ettore Randazzo}
\author[1]{Sami Boukortt}
\author[2]{Alessio Basti}
\author[1]{Mayalen Etcheverry}
\author[1]{Rif A. Saurous}
\author[1]{Ben Laurie}
\author[1]{James Manyika}
\author[1]{Blaise Agüera y Arcas}
\author[1,3,4,5]{Blake A. Richards}

\affil[*]{Equal contributions}
\affil[1]{Google, Paradigms of Intelligence Team}
\affil[2]{HPC Lab, InGeo Department, “G. D’Annunzio” University Chieti-Pescara}
\affil[3]{Mila - Quebec AI Institute}
\affil[4]{McGill University}
\affil[5]{CIFAR}

\begin{abstract}
While traditional evolutionary algorithms hard-code reproduction, self-replication can emerge spontaneously within digital ``primordial soups''. This paper investigates the coevolution of such emergent self-replication alongside problem-solving capabilities. We initialize a population of random 32-byte Z80 assembly programs, requiring self-replication to arise purely through random assembly-level mutations and pairwise program interactions. To couple computation with reproduction, we introduce a task-based validation step: correctly evaluating a polynomial raises a program's interaction probability above a baseline rate. Our experiments yield four primary findings. First, self-replication and mathematical problem-solving successfully coevolve from initial randomness. Second, the pressure to compute accelerates the emergence of compact, robust reproductive architectures that preserve memory for task execution. Third, applying metabolic constraints that penalize runtime promotes the emergence of sophisticated conditional execution patterns that reduce energy use. Finally, partitioning programs into interconnected task niches generates an emergent learning curriculum that utilizes simple solutions as stepping stones toward more complex tasks. Altogether, these results demonstrate an interactive feedback loop: environmental task demands actively shape the physical architecture of self-replication, while spontaneous replication alters the evolutionary trajectory of functional problem-solving.
\end{abstract}

\begin{document}

\maketitle

\section{Introduction}

Origin-of-life research has long debated the fundamental drivers required for life to emerge from non-living matter \citep{life10030020}. Genetics-first accounts hold that life began with self-replicating, information-bearing molecules \citep{Gilbert1986}, whereas metabolism-first accounts posit that energy-harnessing autocatalytic networks preceded genetic replication \citep{Wachtershauser1997-wu, Lancet2018-lm, KAUFFMAN19861, Hordijk2018-gn, kauffman2000investigations}. A complementary tradition treats the two as jointly necessary, with frameworks such as G\'anti's chemoton \citep{ganti2003principles} and Dyson's double-origin hypothesis \citep{dyson1999origins} positing that a minimal living system must couple metabolism with replication. This inquiry extends naturally into Artificial Life (ALife), which seeks to explore the universal principles of living systems across abstract computational substrates \citep{Scharf2015-vx, Fontana1990AlgorithmicCA, hutton2002, kruszewski2022}. Across both biological and digital domains, the spontaneous emergence of self-replication marks a critical phase transition from pre-evolutionary interactions to an evolutionary regime \citep{nowak2008, Aguera2024, Spiegelman1965-ej}. However, platforms studying the evolution of complex behaviors---whether via cellular automata, agentic simulators, or assembly-language organisms---typically bypass the stochastic hurdles of this transition by hard-coding reproduction or seeding their environments with hand-crafted ancestral replicators \citep{neumann1966ca, LANGTON1984135, Sayama1999-ld, oros2007, Schmickl2016-nc, mordvintsev2020growing, Sinapayen2023, chang2018neural, randazzo2021sr, Goldberg1989, randazzo2023biomaker, lu2024jaxlifeopenendedagenticsimulator, Lenski2003, Ofria2004, avida2025, Ray1991, Ray1994}.

To bridge the gap between pre-life emergence and life-era evolution, it is vital to study environments where reproduction is not an axiomatic guarantee. While early ALife models \citep{RASMUSSEN1990111, rasmussen1991matter, PARGELLIS1996111} and recent studies \citep{Aguera2024} have successfully demonstrated that self-replicating programs can spontaneously emerge from random, unseeded initializations without explicit fitness landscapes, these works generally do not address how these nascent self-replication mechanisms might couple with environmental pressures to evolve complex problem-solving. In this paper, we build directly upon the foundational digital primordial soup framework of \cite{Aguera2024} to link the spontaneous emergence of reproduction to the evolution of complex algorithmic behaviors. Specifically, we investigate how the capacity to solve increasingly complex mathematical tasks coevolves alongside the spontaneous origination of reproduction itself.

To explore this, we simulate a population of randomly initialized Z80 assembly programs (executed by an emulated microprocessor). As demonstrated by \cite{Aguera2024}, self-replication can spontaneously emerge from such a primordial soup of random instructions without explicit fitness goals, a result they confirmed in both a minimalist language (BFF) and the Z80 architecture. We adopt Z80 because it is more expressive than BFF while crucially lacking a native multiplication instruction, requiring the evolution of complex iterative loops to perform polynomial evaluation, thus presenting a rugged algorithmic search space. Accordingly, reproduction in our system is not a built-in system-level command, but an emergent behavior that must be performed through the execution of assembly instructions during pairwise program interactions. To investigate the relationship between this primordial reproduction and complex problem-solving, we introduce a competence-modulated interaction rule. Under this regime, a program's opportunity to interact with another and potentially reproduce is contingent upon a validation step: the program is challenged to evaluate a specific mathematical polynomial. This setup differs significantly from platforms such as Avida \citep{Lenski2003, Ofria2004, avida2025}, where solving a task can be viewed as a form of computational metabolism \citep{adami1998introduction} that is rewarded with additional processor time, accelerating execution of the built-in copy instruction, thereby explicitly binding competence and reproduction together. Crucially, our system instead disconnects the two: a successful polynomial evaluation merely increases the likelihood that a program will receive an opportunity to interact with other programs, and does not facilitate or trigger replication. For a lineage to persist in our model, therefore, it must possess its own evolved mechanism for self-replication. Solving the task is thus insufficient for increasing the likelihood of propagation if the program does not simultaneously encode the instructions required to copy itself into the available memory space.

Our experiments yield four primary findings that highlight the interaction between emergent reproduction and the evolution of task solutions. First, we find that self-replication and task-solving coevolve within individual programs. Replication is a strict prerequisite for propagation: a task solution cannot spread through the population unless the program carrying it can also copy itself. Task-solving is not strictly required to reproduce, but it confers a selective advantage, since a program that solves its task is far more likely to be selected for interaction than one that does not. Under this combined pressure, robust reproductive architectures reliably emerge from random noise and, in the same programs, acquire the ability to solve mathematical tasks of increasing complexity. This joint emergence establishes the platform for the three findings that follow.

Second, we find that the pressure to solve tasks reshapes the way the reproductive machinery itself evolves. The earliest replicators to emerge copy themselves by devoting nearly the entire program to the reproduction operation, leaving no room for the instructions needed to solve a task. More compact replication mechanisms occupy just a few bytes and leave the rest of the tape free, so that the same program can both reproduce and solve a task. Because these compact replicators can do both, the demand to solve tasks accelerates the population's transition toward them. In short, the pressure to solve tasks feeds back onto the architecture of how a program reproduces.

Third, we show that applying a metabolic constraint \citep{kempes2017drivers}---decreasing a program's chance to interact based on how many operations it uses to solve a task during the validation step---drives the evolution of conditional operations. Programs evolve to distinguish between the validation step and pairwise interactions, branching their logic to minimize runtime during the task evaluation.

Finally, the structure of reproductive interactions fundamentally shapes the evolution of complex task solutions. Populations evolving in isolation to solve a single complex task consistently fail to progress, even when explicitly rewarded for incremental improvements. This failure exemplifies the \emph{objective paradox}~\citep{Stanley2015, lehman2011abandoning}, where strict objective-driven selection becomes deceptive to evolutionary search. By contrast, complex polynomial solvers emerge reliably only when the population is partitioned into distinct task niches connected by sparse migration \citep{wright1932roles, rainey1998adaptive}. This population structure generates an emergent curriculum, utilizing simpler task solutions as genetic stepping stones. Furthermore, unlike systems with hard-wired reproduction, emergent self-replication supports a significantly wider set of evolutionary paths toward these same solutions.

Altogether, these results demonstrate a bidirectional coupling between emergent self-replication and complex problem-solving. First, spontaneous reproduction acts as the foundational evolutionary operator that enables the persistence and evolution of task solvers. Conversely, the environmental pressure to solve tasks actively shapes the dynamics of reproduction itself, driving emergent replicators toward compact and robust architectures that can seamlessly integrate complex computations.

\section{Results}\label{sec:results}

\subsection{Model overview}\label{subsec:model_overview}

To study the coemergence of self-replication and task solving, we create a population of random programs across a set of separate niches, each assigned a different mathematical task. Evolution proceeds through repeated epochs, in which programs are mutated, tested on their niche's task, and (possibly) paired for interaction. Solving their task makes programs more likely to be selected for interaction. When two programs interact, their combined code is executed, and this execution can have the effect of copying one program's bytes over those of the other. Because an interaction overwrites the two programs with its result, programs that copy themselves tend to spread through the substrate.

\begin{figure}[t!]
    \centering
    \includegraphics[width=0.9\linewidth]{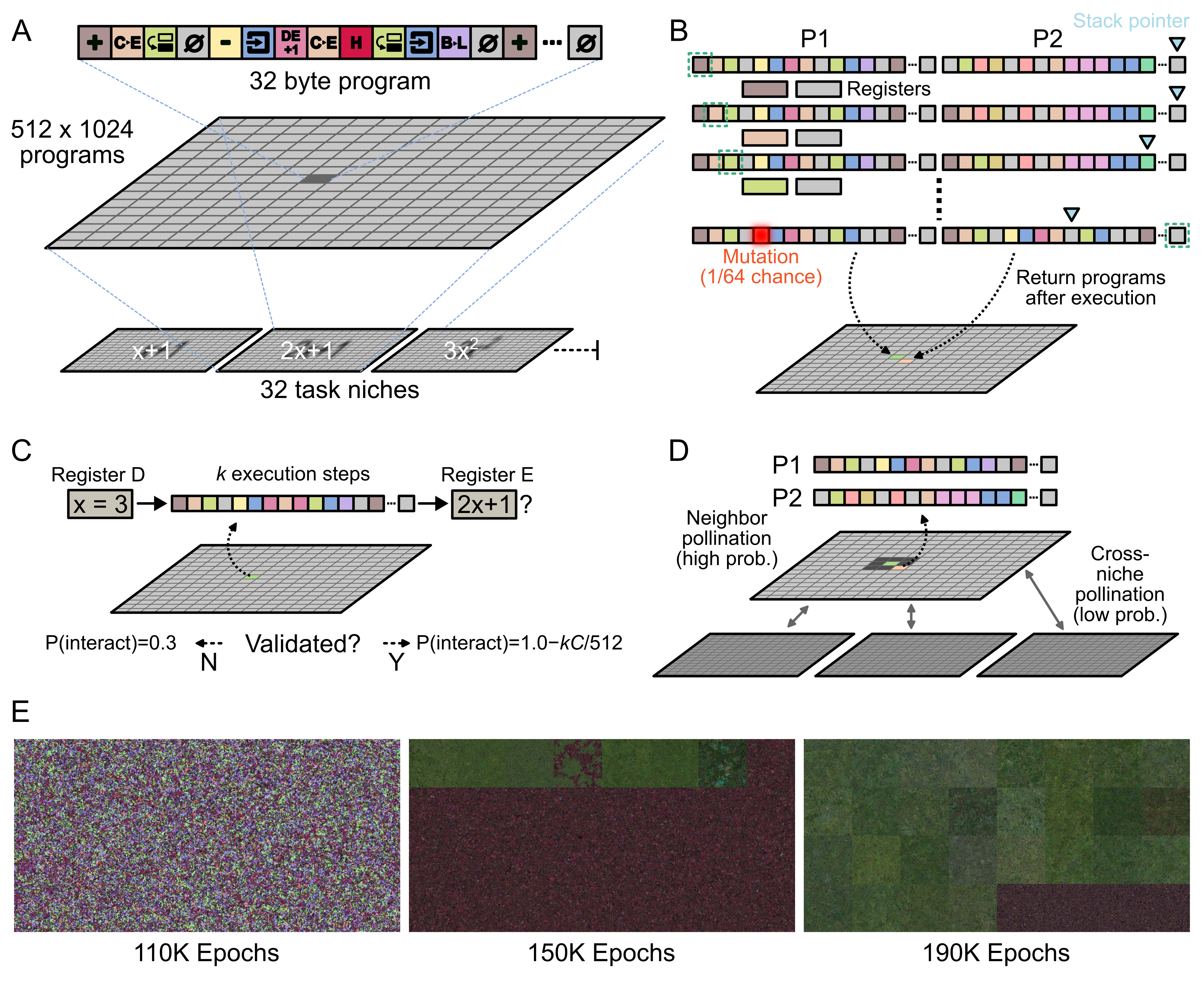}
    \caption{\textbf{System architecture: (A)} The 32 task niches, each a $128 \times 128$ grid of 32-byte Z80 programs and each assigned a target polynomial $f(x)$. \textbf{(B)} Selected pairs concatenate their tapes, execute from left to right, and the resulting halves replace the originals; independently, each program has a 1/64 chance per epoch of one byte being reinitialized. \textbf{(C)} Competence modulation: register \texttt{D} is initialized to a sampled input $x \in \{0,\ldots,15\}$, the program is executed for up to $B=512$ instructions, and register \texttt{E} is read as the output. Producing $f(x)$ on three distinct inputs raises the interaction probability from $p_{\mathrm{base}}=0.3$ to $p_{\mathrm{succ}}=1$, optionally discounted by a metabolic penalty proportional to execution steps. \textbf{(D)} Most pairings are local von Neumann neighbor interactions; with probability $\pi$ the partner is instead drawn uniformly from any niche (cross-niche pollination, CNP). \textbf{(E)} A view of three intermediate states of the system at different epochs of a simulation. Each pixel in the grid corresponds to a program. Each program's color is obtained by averaging the RGB vectors assigned to a subset of its bytes according to a handcrafted colormap. The color of programs that successfully passed their most recent validation is biased toward green. Note the homogenization of the program content as self-replicators take over, as well as the gradual emergence of solutions in specific niches (each block corresponds to a different niche).}
    \label{fig:system}
\end{figure}

Concretely, the population is divided into 32 niches, and each niche is an isolated two-dimensional grid of cells, with one program per cell. A program is a sequence of $\ell = 32$ bytes, which we also call a ``tape'' (Fig.~\ref{fig:system}A). Within a niche, a program interacts mostly with its immediate neighbors on the grid. The niches have no spatial arrangement relative to one another, so they act as a collection of separate populations. Each niche is assigned one of 32 polynomial functions, and a program's task is to evaluate the assigned polynomial. Across all niches the population totals $2^{19}$ programs, all initialized as random byte sequences.

Each epoch begins with a mutation step, in which every program has a small chance of having one random byte reinitialized (Fig.~\ref{fig:system}B). The programs are then evaluated, and a subset is selected for pairwise interaction, as we describe next.

The program evaluation is a competence test we call ``validation''. The program is run on its own by a Z80 emulator (see \nameref{sec:methods} for details), and we check whether it evaluates its niche's polynomial correctly on a few randomly sampled inputs. This validation yields a binary success or failure signal: the program passes only if it computes the correct output for all sampled inputs. At the end of validation, the final state is written back to the cell, so any modification the program makes to its own bytes during validation persists.

A program that passes validation becomes much more likely to be selected for interaction, though this likelihood is reduced by a metabolic penalty that grows with the number of steps it used to pass the validation (Fig.~\ref{fig:system}C). A program that fails is still selected sometimes, but only at a low baseline rate (see \nameref{sec:methods} for the exact rule).

A selected program is then paired with a partner. Most of the time, with probability $0.95$, the partner is one of its immediate neighbors in the same niche. The rest of the time the partner is any program drawn uniformly from the whole population (ignoring niche membership, see Fig.~\ref{fig:system}D). This occasional long-range pairing, which we call cross-niche pollination (CNP), lets a solution found in one niche spread into another, and we show later that it is key for solving the harder tasks.

In an interaction, the two programs' tapes are concatenated into a single $2\ell$-byte memory space, and execution begins at the first byte of the first program. Whatever the combined tape contains at the end of a fixed instruction budget is split back into two 32-byte halves and written back into the cells the two programs came from, carrying the result into the next epoch. Replication is never imposed by the system; it occurs only when the executed code, on its own, copies one program's bytes over the second program's memory region.

%%%%%%%%%%%%%%%%%%%%%%%%%%%%%%%%%%%%%%
%%%%%%%%%%%%%%%%%%%%%%%%%%%%%%%%%%%%%%

\subsection{Self-replication and task solutions emerge from a primordial soup}

Over one million epochs of evolution, robust self-replication and task-solving capabilities consistently coemerged from completely random initial programs, as successful self-replicating lineages spread to dominate the grids (Fig.~\ref{fig:system}E). This result shows that spontaneous replication and the ability to solve increasingly complex tasks can coevolve. Indeed, the two abilities reinforce each other. A program that cannot replicate is inevitably overwritten by one that can, regardless of its task-solving competence. Conversely, among replicating programs, those that successfully solve their niche's polynomial are selected for interaction at a higher rate. Over time, the lineages that dominate a niche tend to integrate both behaviors (see Supplementary Information Sec.~\ref{app:discovered_architectures} for representative annotated program listings across linear, quadratic, and cubic tasks).

In contrast to a standard evolutionary algorithm, our system never copies or reproduces programs. It provides only random initialization, mutation, and a validation step that biases how frequently programs interact. This bias induces a fitness landscape, but the system provides no built-in means of navigating it, since reproduction must be discovered by the executed code. The evolutionary machinery therefore emerges from random bytes together with the ability to solve problems, rather than being supplied in advance.

\subsection{Tasks alter the dynamics of emergent self-replication}
\begin{figure}[t!]
    \centering
    \includegraphics[width=0.75\linewidth]{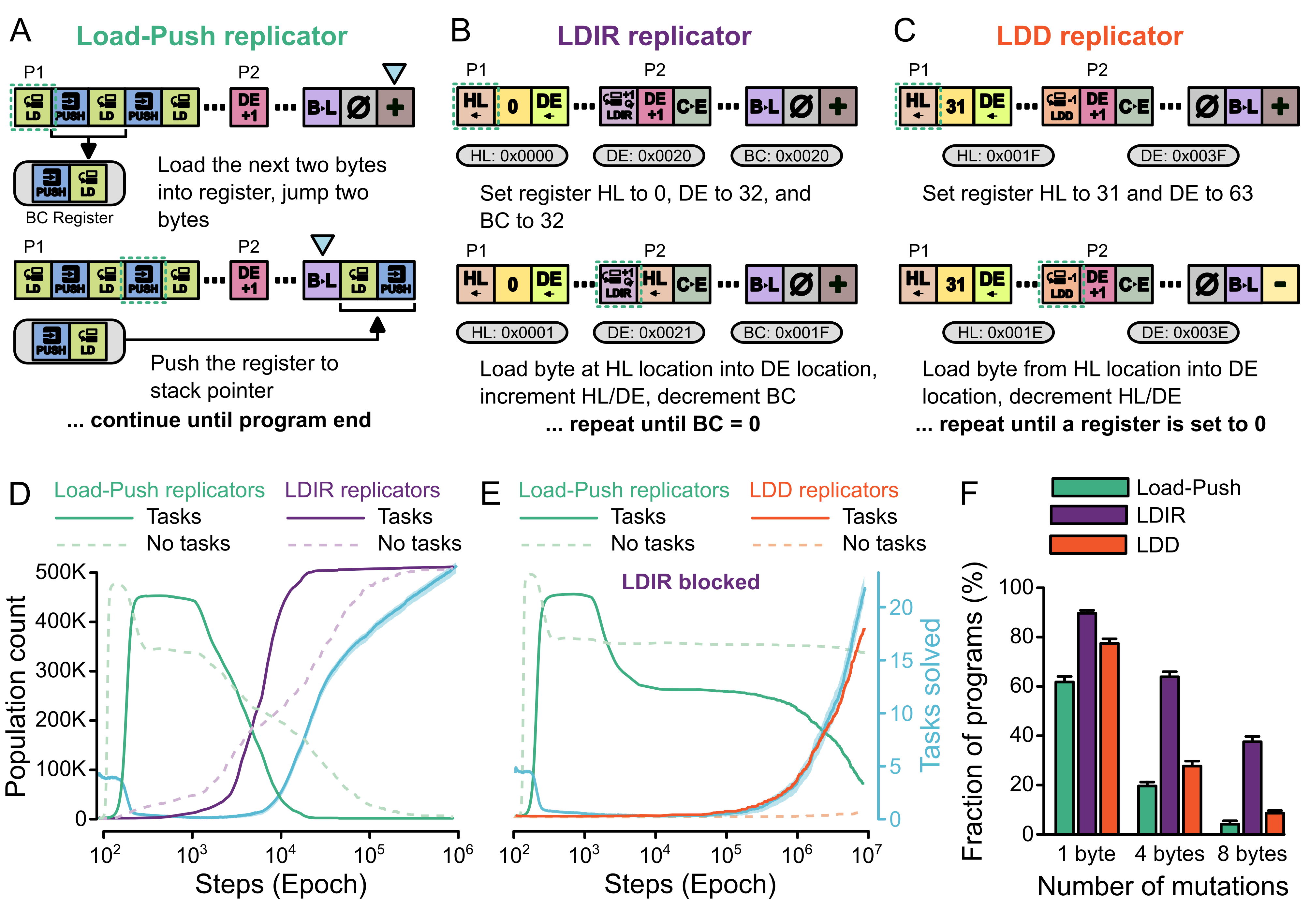}
    \caption{\textbf{Emergence of self-replicators and accelerating effect of tasks: (A)} Illustration of a Load--Push replicator. \textbf{(B)} Illustration of an LDIR replicator. \textbf{(C)} Illustration of an LDD replicator. \textbf{(D)} Populations of Load--Push and LDIR replicators over time, with and without tasks. Data shown is mean over 100 seeds. Number of tasks solved also shown as mean $\pm$ 95\% C.I. \textbf{(E)} Populations of Load--Push and LDD replicators over time with and without tasks, when LDIR replicators are blocked. Data shown is mean over 100 seeds. Number of tasks solved also shown as mean $\pm$ 95\% C.I. \textbf{(F)} Percentage of programs with no replication errors following one byte mutation (\textit{left}), four byte mutations (\textit{middle}), and eight byte mutations (\textit{right}) for Load--Push, LDIR, and LDD replicators. Error bars represent 95\% Wilson score intervals. LDIR replicators were significantly more robust than LDD ones, which in turn were significantly more robust than Load--Push across all mutation levels (one-sided two-proportion $Z$-test with Bonferroni correction; $n = 100$ per group; all pair-wise comparisons showed $p < 0.05$ after correction, max adjusted $p = 0.042$ for LDD vs. Load--Push at 8 mutations).}
    \label{fig:effects_of_tasks}
\end{figure}

The coemergence of replication and task-solving raises a natural question: how do the two interact as evolution proceeds? One might expect the dynamics of replication to be unaffected by the tasks, with task pressure shaping only the task-solving code that programs carry. Instead, we found a more intricate coupling, in which the pressure to solve tasks and the architecture of replication shape one another. Two types of replicators came to dominate the population at different epochs. The first, a simple replicator we call `Load--Push', consists of a string of paired `Load' and `Push' instruction bytes. Because the Z80's \texttt{PUSH} instruction operates on 16-bit registers, each pair copies the program two bytes at a time to the adjacent memory space of a partner program (Fig.~\ref{fig:effects_of_tasks}A). Lacking a loop, full replication requires a long chain of these pairs, meaning a Load--Push replicator consumes the entire 32-byte tape.

A second replicator, the `Load Increment Repeat' (LDIR) replicator, reliably took over later. It utilizes a specialized Z80 block-copy instruction (\texttt{LDIR}) that automatically loops to copy a sequence of bytes from one location in memory to another. Rather than a long, repetitive sequence of instructions, it needs only to initialize its copy parameters to copy from the executing program (P1) into its partner (P2) once before the single instruction executes the entire copy. This allows a program to achieve replication using only a few bytes (Fig.~\ref{fig:effects_of_tasks}B), leaving the rest of its memory free for task-solving code (see \nameref{sec:methods} for the exact register mappings).

As noted above, the Load--Push replicators were the first to appear and dominate the grids (Fig.~\ref{fig:effects_of_tasks}D, \textit{solid green line}), but they eventually gave way to the LDIR replicators (Fig.~\ref{fig:effects_of_tasks}D, \textit{solid purple line}). The number of tasks solved tracked this turnover (Fig.~\ref{fig:effects_of_tasks}D, \textit{solid blue line}). Specifically, the number of tasks solved \textit{dropped} as the Load--Push replicators spread, then recovered and grew as the LDIR replicators took over. The reason for this dynamic is that a Load--Push replicator uses the entire tape to copy itself, leaving no room for task-solving code, so its spread displaces programs that solve tasks. The LDIR replicator, by contrast, needs only a few bytes and leaves the rest of the program free; accordingly, its emergence always preceded the growth in task-solving programs, indicating that it unlocked the ability to both replicate and solve tasks. The takeover does not depend on the tasks: LDIR replicators are more robust to mutation than Load--Push replicators (see Fig.~\ref{fig:effects_of_tasks}F), and removing the task-based gating of interactions (see \nameref{sec:methods} for details) slowed the Load--Push-to-LDIR transition but did not prevent it (Fig.~\ref{fig:effects_of_tasks}D, \textit{dashed lines}). Task pressure nonetheless speeds it up considerably, because the free tape of an LDIR replicator can accumulate task-solving code, and a program that both replicates and passes validation is selected for interaction more often, which accelerates its spread.

To determine whether the accelerating effect of task pressure is specific to the \texttt{LDIR} instruction, we ran control experiments where we removed \texttt{LDIR} and its close relatives (\texttt{LDI} and \texttt{LDDR}) from the language. In their absence, we observe the consistent emergence of a different replication mechanism based on the \texttt{LDD} instruction, though it takes over the population much more slowly than LDIR had (Fig.~\ref{fig:effects_of_tasks}E, \textit{solid lines}). This transition occurs consistently only under task pressure; without it, the transition from Load--Push is extremely slow and fails to complete within ten million epochs (Fig.~\ref{fig:effects_of_tasks}E, \textit{dashed lines}). While \texttt{LDIR} has a built-in repeat mechanism that copies the entire program once its parameters are initialized, \texttt{LDD} copies only a single byte at a time. An \texttt{LDD}-based replicator must therefore not only initialize its parameters to copy from P1 to P2 but also implement an explicit, multi-byte loop to achieve replication (Fig.~\ref{fig:effects_of_tasks}C), representing a more complex reproductive architecture. These results show that when one solution for reproduction is blocked, another is found, and that task pressure again accelerates the change.

To verify our hypothesis that LDIR replicators naturally come to dominate because there is a robustness hierarchy of the replicators, we quantified how well each reproduction architecture tolerates replication errors (see \nameref{sec:methods}). Our measurements revealed a clear robustness hierarchy (LDIR $>$ LDD $>$ Load--Push), where LDIR programs remain functional after successive mutations significantly more often than LDD loops, which in turn outperform the Load--Push sequences (Fig.~\ref{fig:effects_of_tasks}F).

Taken together, these experiments show that the way a program reproduces keeps evolving after self-replication first appears. Robustness to mutation drives the population toward more compact replicators on its own, but the pressure to solve mathematical tasks accelerates that shift, because a compact replicator leaves room in the tape for task-solving code. The fitness landscape created by the tasks thus reshapes the replication machinery, so that how a program computes and how it reproduces evolve together.

%%%%%%%%%%%%%%%%%%%%%%%%%%%%%%%%%%%%%%
%%%%%%%%%%%%%%%%%%%%%%%%%%%%%%%%%%%%%%

\subsection{Metabolic constraints drive efficient, conditional execution}\label{subsec:metabolism}

In biological systems, computation is not free; organisms must balance the energy spent processing information with the resources required for replication. To study how such efficiency constraints shape algorithmic evolution, we introduced a ``metabolic cost'' to the validation phase, discounting a program's interaction probability in proportion to the number of execution steps it takes to solve its task (see \nameref{sec:methods}). We asked whether this constraint would drive the evolution of more efficient programs, and how it might affect the replication phase where the constraint does not apply.

\begin{figure}[t!]
    \centering
    \includegraphics[width=0.75\linewidth]{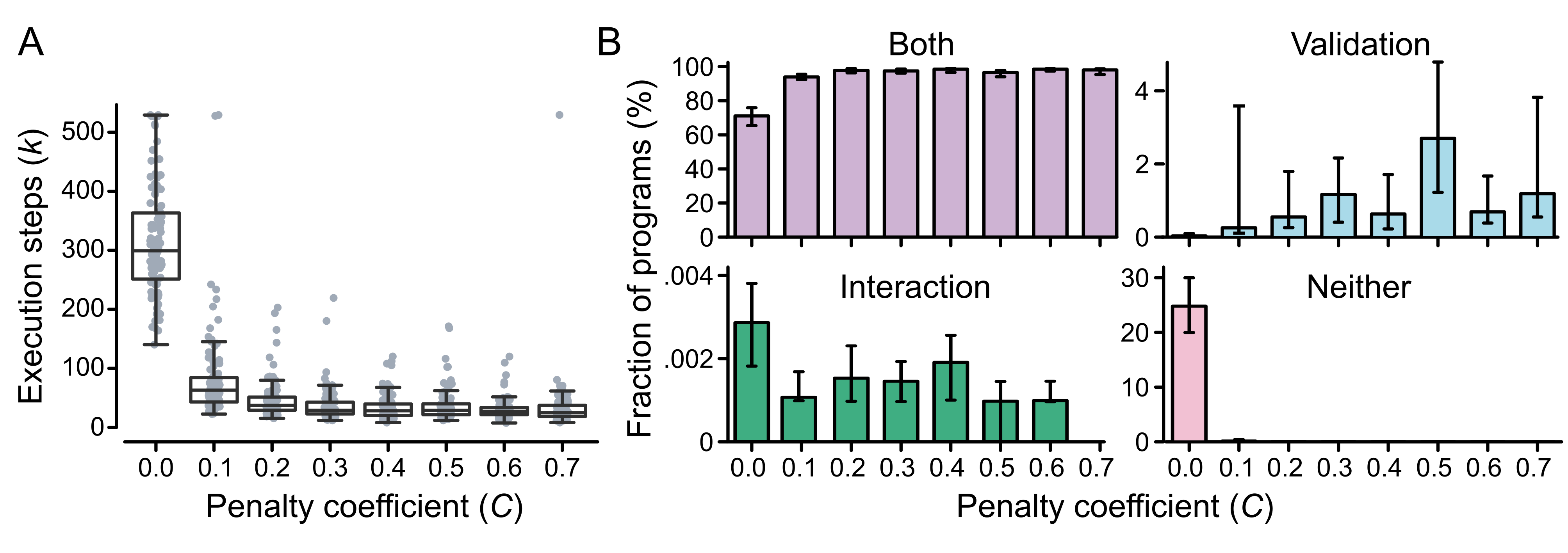}
    \caption{\textbf{Impact of metabolic constraints on program halting behavior: (A)} Average number of execution steps of a run as a function of the strength of the metabolic penalty coefficient. Data points show values for individual runs (100 runs for each of the 8 penalty coefficients). Box plots show the median and interquartile range (IQR), with whiskers extending to $1.5 \times \text{IQR}$. The penalty coefficient correlates negatively with the average execution length (Spearman's rank correlation; $\rho = -0.61$, $p < 0.0001$, $N = 800$). \textbf{(B)} Percentage of programs with different forms of halting behavior. Data shown is median with bootstrapped $95\%$ C.I. over 100 simulations with different random seeds for each of the 8 penalty coefficients. The occurrence of conditional halting at the validation phase positively correlated with the metabolic penalty coefficient (Spearman's rank correlation; $\rho = 0.21$, $p < 0.0001$, $N = 800$).}
    \label{fig:halting_behavior}
\end{figure}

While the metabolic constraint left the overall rate of task solution emergence unchanged, it profoundly altered how those solutions were implemented. As the metabolic penalty coefficient increased, the average number of execution steps spent during validation fell significantly (Fig.~\ref{fig:halting_behavior}A). This efficiency was achieved because evolved programs consistently began incorporating a \texttt{HALT} instruction, which terminates execution as soon as the correct output is computed.

In a minority of simulations, programs went further, evolving context-dependent behavior by exapting register \texttt{D} as a sensory cue to distinguish between the validation and interaction phases. In our execution protocol, the emulated environment uses register \texttt{D} to pass the task input \(x \in \{0,\ldots,15\}\) during validation, meaning it contains a non-zero value fifteen times out of sixteen. During the interaction phase, however, the environment always initializes \texttt{D} to zero. Evolving programs occasionally co-opted this difference, executing a conditional branch that triggers a \texttt{HALT} instruction only when the value in \texttt{D} is non-zero. Although this conditional halting remained a minority strategy in absolute terms, the probability of its emergence rose significantly as the metabolic penalty became more severe (Fig.~\ref{fig:halting_behavior}B; see also Supplementary Information Sec.~\ref{app:discovered_architectures} for examples of lineages that similarly leverage register \texttt{D} to skip the LDIR replication preamble during validation).

This behavior represents a case of one evolved sequence serving two distinct roles depending on the environmental context. Because the same program is executed during both validation and interaction, the metabolic pressure on computation (the task-solving phase) can shape how the program behaves during reproduction (the interaction phase), even though the reproductive step itself carries no metabolic penalty. This provides a further example of how computation and reproduction can become coupled within a single, short program.

%%%%%%%%%%%%%%%%%%%%%%%%%%%%%%%%%%%%%%
%%%%%%%%%%%%%%%%%%%%%%%%%%%%%%%%%%%%%%

\subsection{Cross-niche pollination produces emergent curricula}\label{subsec:cnp}

To evaluate the learning capabilities of a system where reproduction and problem-solving coevolve, we tested its capacity to solve tasks of increasing complexity. In our system, solving higher-order polynomials depended crucially on the combination of a multi-task environment and sparse connectivity between its task niches. Because niches exchanged material sparsely through cross-niche pollination, solutions to simpler tasks could seed niches working on harder ones. This spatial structure allowed the population to advance through a sequence of intermediate problems, forming an emergent curriculum that led toward a complex target. While such curricula are typically hand-designed by an experimenter to guide evolutionary search, in our system they arose spontaneously from the niche structure and the sparse pollination that propagated solutions between niches.

\begin{figure}[t!]
    \centering
    \includegraphics[width=0.65\linewidth]{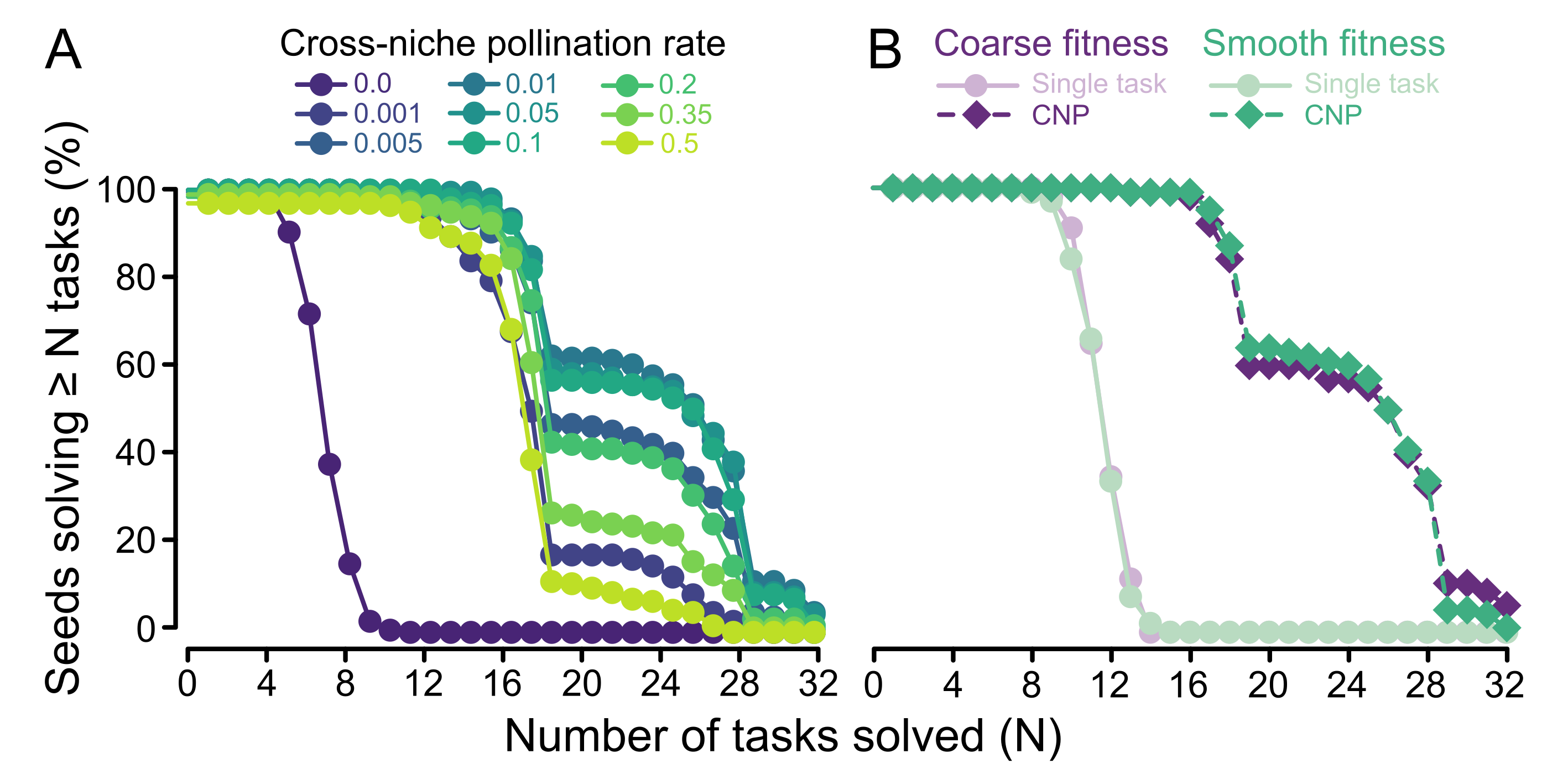}
% TODO: Panel A reports n = 200 per group but Table 1 says 100 replicates per condition; confirm and state where the 200 come from.
    \caption{\textbf{Impact of niche dynamics on task success: (A)} Percentage of simulations where at least $N$ tasks are solved for different rates of cross-niche pollination (CNP). A CNP rate of 0.05 solved significantly more tasks compared to both the 0.0 and 0.5 rates (Mann--Whitney $U$ test; vs. 0.0: $U = 0.5$, $p < 0.0001$; vs. 0.5: $U = 6858.0$, $p < 0.0001$; $n = 200$ per group). \textbf{(B)} Percentage of simulations where at least $N$ tasks are solved when there are niches (CNP) or not (Single Task), and when the fitness function is smoothed or not. CNP significantly increased the number of tasks solved compared to single-task setups under both coarse ($U = 9979.0$, $p < 0.0001$) and smooth ($U = 9976.5$, $p < 0.0001$) fitness landscapes. However, there was no significant difference in performance between smooth and coarse fitness in either setup (Mann--Whitney $U$ test; all $p > 0.65$; $n = 100$ per group).}
    \label{fig:niche_behavior}
\end{figure}

The effectiveness of this emergent curriculum depends critically on the rate of program interaction between niches. Varying the cross-niche pollination (CNP) rate (the probability that a program's interaction partner is drawn uniformly from the whole population rather than from its local neighborhood; see \nameref{sec:methods}) revealed that moderate rates of CNP were essential for solving higher-order tasks (Fig.~\ref{fig:niche_behavior}A). At one extreme, fully segregated niches (a rate of zero) solved nothing beyond first-order polynomials, while at the other extreme, high rates impaired the evolution of solutions to higher-order polynomials.

%%%%%%%%%%%%%%%%%%%%%%%%%%%%%%%%%%%%%%
%%%%%%%%%%%%%%%%%%%%%%%%%%%%%%%%%%%%%%

We tested whether explicitly smoothing the fitness function by other means could replicate the benefit of applying a moderate CNP rate. The binary success/fail signal was therefore replaced with a graded one, scaling the interaction probability linearly with how far a program's outputs fell from the correct values on the validation inputs (see \nameref{sec:methods} for the exact formulation). We ran a control setup where each of the tasks was evolved in complete isolation (the `Single Task' setup). In these runs, we kept the exact same overall compute and memory budget as the CNP system. Consequently, the population dedicated to a single task ($524{,}288$ programs) was organized in a $512 \times 1024$ grid, 32 times larger than a single $128 \times 128$ niche used in the CNP experiment. Despite this 32-fold population advantage per task, these isolated populations failed to progress, solving at most 15 of the simplest tasks out of the total set (Fig.~\ref{fig:niche_behavior}B). In contrast, the CNP system, which partitioned the same global population across 32 smaller, interconnected niches, drastically outperformed them. This difference reveals a dynamic related to the objective paradox within algorithmic evolution---the phenomenon where directly optimizing for a complex target objective often prevents its discovery, whereas optimizing for intermediate stepping stones succeeds. Assembling complex logic (such as nested accumulation loops for polynomial evaluation; Supplementary Information Sec.~\ref{app:discovered_architectures}) requires distinct, discontinuous algorithmic modules, and the niche structure of the CNP system allowed these modules to be combined across niches, whereas they would be unlikely to arise in a single-task population.

\begin{figure}[t!]
    \centering
    \includegraphics[width=0.95\linewidth]{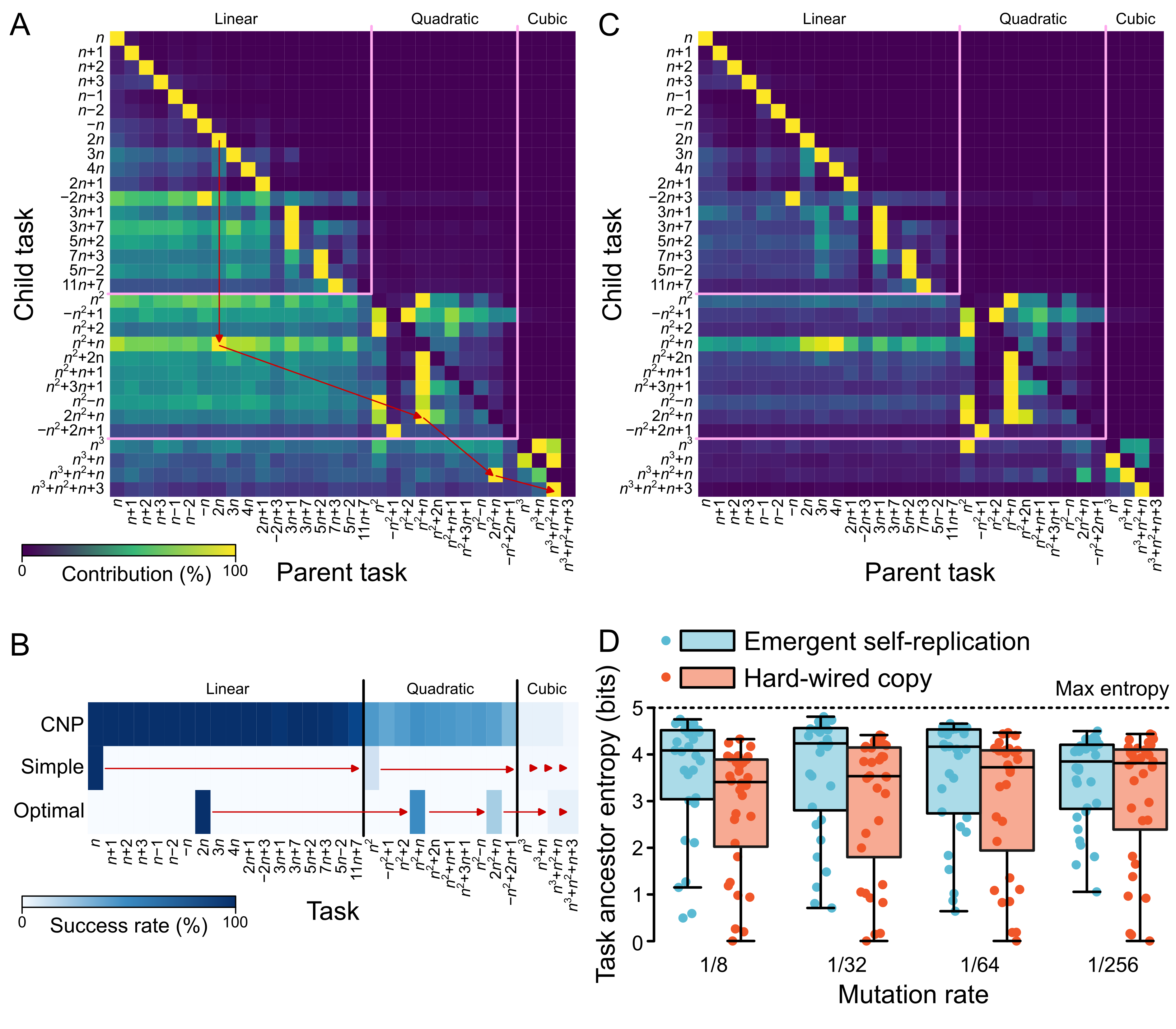}
% TODO: Panel B CNP row appears uniformly dark through the cubic tasks, yet Fig. 6 reports 4-7% cubic solve rates; check the rendering. Also relabel the panel A colorbar from "Contribution (%)" to normalized contribution.
    \caption{\textbf{Evidence of an emergent curriculum: (A)} Genealogy of programs based on most recent ancestors' niches. Note the non-uniform, non-symmetric genealogy, implying that solutions to certain tasks tend to evolve from existing solutions to specific easier tasks. Red arrows indicate a common genealogical path, used for constructing the `optimal' curriculum. Colors show ancestral contributions, min--max normalized within each row to $[0,1]$ (see \nameref{sec:methods}); the colorbar shows this normalized value rescaled to 0--100. \textbf{(B)} Success rate per task for the full cross-niche pollination system (\textit{top}), a simple curriculum moving from easier to harder tasks (\textit{middle}), and a curriculum based on the genealogy shown by the red arrows in panel \textbf{A} (\textit{bottom}). Note the greater success on quadratic and cubic tasks for the optimal curriculum relative to the simple curriculum. \textbf{(C)} Genealogy of programs in the modified system with hard-wired reproduction. Note the narrower ancestor distribution compared to panel \textbf{A}, indicating a less diverse set of evolutionary paths toward the same tasks. \textbf{(D)} Entropy of the ancestor niche distribution across a sweep of per-program mutation rates for both systems. The system with emergent self-replication maintains a broader ancestor distribution across all tested rates. Notably, even when comparing the mutation rate where emergent self-replication achieves its lowest median entropy against the rate where hard-wired reproduction achieves its highest median entropy, the emergent system maintains significantly higher entropy across tasks (paired Wilcoxon signed-rank test; $W = 512.0$, $p = 0.0007$, $n = 32$ tasks).}
    \label{fig:curriculum_dynamics}
\end{figure}

To trace how solutions spread across these niches, we tracked the lineage of each program by recording the solved niche from which it received material most recently (see \nameref{sec:methods}). Aggregating these ancestral contributions reveals a structured, non-symmetric genealogy across tasks (Fig.~\ref{fig:curriculum_dynamics}A), showing that solutions to higher-order polynomials systematically descend from specific intermediate niches.

To test whether this genealogy identifies useful stepping stones, we compared the full cross-niche pollination system against two explicit curricula, in which the entire population is evaluated on a single task that advances once solved (see \nameref{sec:methods}): a simple curriculum that orders tasks from easier to harder, and an `optimal' curriculum that follows the common genealogical path marked by the red arrows in Fig.~\ref{fig:curriculum_dynamics}A. The optimal curriculum performed markedly better on quadratic and cubic tasks than the simple curriculum (Fig.~\ref{fig:curriculum_dynamics}B).

%%%%%%%%%%%%%%%%%%%%%%%%%%%%%%%%%%%%%%
%%%%%%%%%%%%%%%%%%%%%%%%%%%%%%%%%%%%%%

\subsection{Emergent self-replication alters the curriculum structure}\label{subsec:curriculum_structure}

As shown in Fig.~\ref{fig:curriculum_dynamics}A, for a given target task, the specific parent niche that seeds its solution varies across independent runs. To ask whether this genealogical breadth depends on the endogenously evolved replication mechanism, we evaluated a control system in which reproduction is hard-wired. In this variant, the standard interaction between paired programs is replaced with a process that directly overwrites the second program with an exact copy of the first without executing Z80 code, and memory writes during validation are discarded. In this setting, programs no longer need to maintain their own self-replication logic within the 32-byte tape, making random mutation the sole source of heritable change. All other parameters, including niche topology, cross-niche pollination rate, and validation rules, remained identical.

We traced lineages in both systems with the ancestor-tracking procedure (see \nameref{sec:methods}), labeling each program with the most recent solved niche from which its lineage received material. Every program starts labeled with its own niche and keeps that label until an interaction transfers a new label from its partner (see \nameref{sec:methods} for the exact update rules). Such a label is not the same as identifying the niche that contributed the code responsible for the solution. A lineage can pass through a solved niche, inherit its label, and later solve its own task through code that owes little to that passage: the solving code may descend from ancestors earlier than the labeled niche, or arise through mutations after the label was set. Because of such spurious ancestors, the ancestor contribution matrices (Fig.~\ref{fig:curriculum_dynamics}A, C) are best read as showing which solved niches seeded replicating material into each target niche, rather than as identifying where each solution's code originated.

Read this way, both systems produce structured, non-symmetric genealogies, which confirms that the emergent curriculum follows from the niche structure and sparse pollination rather than from how replication is implemented. But, despite these similarities, the two replication mechanisms differ in how concentrated those genealogies are. In the modified system with hard-wired replication, the ancestor distribution of each task is considerably narrower (Fig.~\ref{fig:curriculum_dynamics}C): a given task is reached from the same parent niche in most runs. In contrast, when self-replication must evolve, the same task is reached from several different ancestors. We quantified this difference in the entropy analysis in Fig.~\ref{fig:curriculum_dynamics}D. Emergent self-replication therefore supports a wider set of evolutionary paths toward the same solution.

One might expect this difference to merely reflect a higher rate of genomic change in the default system, since executing the concatenated tape gives a single mutated byte the opportunity to disrupt the copy operation and introduce further changes in the offspring, while the modified system copies the parent exactly. To test whether noise alone accounts for the broader genealogies, we recomputed the ancestor contribution matrices over a sweep of per-program mutation rates in both systems. The system with emergent self-replication retained the broader ancestor distribution at every rate we tested (measured with the entropy of the ancestor niche distribution), including against modified-system runs with mutation rates well above the original rate (Fig.~\ref{fig:curriculum_dynamics}D). The extra genealogical variance is thus not simply a byproduct of additional noise, but of replication being an evolving property of the programs themselves.

This diversity of paths is not itself an advantage in solving tasks, though. With the whole tape available for computation and inheritance guaranteed to be faithful, the modified system solves more of the harder tasks than the emergent self-replication system does (Supplementary Information Sec.~\ref{app:control_performance}). Emergent replication acts here on the structure of the curriculum: when a population must evolve how it reproduces, the same task tends to be reached by more varied evolutionary routes. This suggests that the coevolutionary setup does not simply tend to approximate standard evolutionary search, but constitutes a qualitatively distinct search process which comes with a broader sampling of the routes that connect one solution to the next.

\section{Discussion}\label{sec:discussion}

In this paper, we examined the coevolution of self-replication and problem solving within a digital primordial soup (Fig.~\ref{fig:system}). Building upon \cite{Aguera2024}, we found that self-replication and task solutions both reliably emerged from random noise. Unlike traditional evolutionary systems where reproduction is a fixed background rule, here replication and problem-solving mechanisms coevolved and shared a 32-byte memory. The transition from pre-evolutionary stochasticity to a coevolutionary regime was shaped by the pressure exerted by the tasks. This computational pressure drives a feedback loop where the demand to solve tasks actively reshapes the machinery of heredity itself, steering populations away from naive copying mechanisms that monopolize the tape and accelerating the emergence of more compact, robust replication architectures, such as \texttt{LDIR} and \texttt{LDD} loops, that leave room for task-solving code (Fig.~\ref{fig:effects_of_tasks}). In this way, the demands placed on what a program computes feed back onto the architecture of how it reproduces. Similarly, metabolic constraints encouraged the evolution of programs that halt at appropriate times to optimize their execution budget. Notably, these halting behaviors sometimes evolved to function in a conditional manner, allowing programs to distinguish between validation and interaction executions (Fig.~\ref{fig:halting_behavior}).

We leveraged separate task niches connected by sparse pollination as a framework to investigate the learning capabilities of this coevolutionary system. In our experiments, populations working in isolation consistently failed to solve non-trivial polynomials. This failure demonstrates that direct optimization toward a difficult target is fundamentally insufficient in this landscape. Instead, partitioning the population into separate niches connected by sparse pollination allowed the system to self-organize an evolutionary curriculum, where simpler tasks served as essential stepping stones to seed solutions for more complex ones (Fig.~\ref{fig:niche_behavior}). When we compared this self-organized process to hand-designed curricula, a simple path of steadily increasing complexity failed, while a curriculum retracing the emergent genealogy performed markedly better on quadratic and cubic tasks. This contrast highlights that the niche structure acts as a decentralized engine for discovering evolutionary pathways (Fig.~\ref{fig:curriculum_dynamics}).

One limitation of this study is that we did not systematically investigate the interaction dynamics for the emergence of genetic diversification, such as recombination or symbiogenesis. Our interaction protocol does not preclude genetic mixing, and accidental recombination may have played a role in evolutionary transitions. This contrasts with traditional genetic algorithms where crossover is fixed \textit{a priori} \citep{Goldberg1989, eiben2015introduction}. Most of the dominant replicators we manually inspected reproduced asexually, though the genealogical comparison in Sec.~\ref{subsec:curriculum_structure} suggests that this regime is not simply approximating one with faithful asexual copying. Investigating whether programs can spontaneously discover reciprocal genetic exchange remains an important direction for future work.

Another limitation of this study is that the tasks used (calculating polynomials) were, despite Z80 lacking a multiplication instruction, relatively simple and drawn from the same class. While the use of simple tasks and restricted domains is in line with previous work on algorithmic evolution in ALife \citep{Ofria2004, avida2025}, follow-up research could explore how tasks and self-replication coevolve when the tasks and their domains are wider and drawn from a real-world relevant set. Finally, there was a ceiling on the complexity of algorithmic solutions that could be discovered, both due to the hard limit of 32-byte program lengths (though in extended 64-niche environments programs can evolve nested Horner-style factorizations to evaluate polynomials as high as sextic degree; Supplementary Information Sec.~\ref{app:horner_evaluation}), and due to the absence of cooperative mechanisms for programs to co-solve tasks. Future research could explore how self-replication and tasks coevolve when more complex, potentially multi-agent, solutions are possible.

In total, our work demonstrates that mechanisms of inheritance and fitness optimization can coevolve to shape one another, and supports the perspective that origin-of-life and evolutionary computation research can view evolution not merely as a static search algorithm, but as a dynamic, plastic property of an open-ended learning system.

\section{Methods}\label{sec:methods}

\subsection{Competence-modulated interaction and metabolic constraints}\label{subsec:methods_competence}
% TODO: State initialization of IX/IY and the shadow register set (B.1.5/B.1.6 hard-wired listings execute EX AF,AF').
Interactions between programs are biased by their ability to solve the polynomial task assigned to their niche (Fig.~\ref{fig:system}C). Before a selected program \(P_1\) is allowed to interact, it undergoes a validation phase. We place \(P_1\) in the first half of the \(2\ell\)-byte memory and zero the second half, and execute the tape for a maximum budget of \(B=512\) instructions. For this validation run, the Z80 registers are initialized as follows: registers \texttt{HL}, \texttt{BC}, and \texttt{E}, and the program counter \texttt{PC} are set to zero; registers \texttt{A} and \texttt{F}, and the stack pointer, are set to \texttt{0xFF} (so the zero, carry, sign, and parity/overflow flags are all set at program entry); and register \texttt{D} is initialized with the task input \(x\). Since the emulated address space is limited to $2\ell$ bytes and every memory access is taken modulo $2\ell$, the stack pointer value \texttt{0xFF} resolves to the last byte of the tape, so stack writes grow backwards from the end of the memory. The output is read from register \texttt{E} (which, being an 8-bit register, implicitly stores the result modulo $2^8$). During the subsequent interaction phase, the register initialization is identical, except that register \texttt{D} is initialized to zero rather than the task input.

To pass validation, \(P_1\) must satisfy \(\mathtt{E}=f(x) \bmod 2^8\) for three inputs drawn without replacement from \(\{0,\ldots,15\}\), where \(f\) is the polynomial associated with \(P_1\)'s niche. The three validation runs are executed in sequence on the same program: each run re-initializes the registers and the zeroed second half of memory, while any modification the program makes to its own bytes carries over into the next run, and the state after the last executed run is what persists in the cell. Evaluating a subset of three inputs is sufficient to make accidental successes by incorrect programs highly improbable. Programs that pass receive an elevated interaction probability \(p_{\mathrm{succ}}=1.0\); programs that fail keep the baseline probability \(p_{\mathrm{base}}=0.3\).

To simulate metabolic constraints, we discount the interaction probability of validated programs in proportion to the number of execution steps they spend during validation. The interaction probability is
\begin{equation*}
p(\text{interaction}) =
\begin{cases}
p_{\mathrm{succ}} - C \dfrac{k}{B} & \text{if validated}, \\[6pt]
p_{\mathrm{base}} & \text{otherwise},
\end{cases}
\end{equation*}
where \(k\) is the average number of execution steps taken over the three validation runs, \(B=512\) is the maximum instruction budget, \(p_{\mathrm{succ}}=1.0\) and \(p_{\mathrm{base}}=0.3\) are the success and baseline interaction probabilities, and \(C\) is the metabolic penalty coefficient. The penalty is applied only to validated programs; the probability of a failed program is fixed at \(p_{\mathrm{base}}=0.3\) regardless of its execution cost. Unless otherwise stated we set \(C=0.3\); the sweep reported in Fig.~\ref{fig:halting_behavior} instead varies \(C\) across \([0, 0.7]\).

Validation is distinct from solving a task. Validation gates interaction and is decided on three sampled inputs, whereas we report a task as solved on the basis of the full input range. Concretely, throughout the paper we count a niche as having solved its task when at least \(10\%\) of its programs produce \(f(x) \bmod 2^8\) correctly for every \(x \in \{0,\ldots,15\}\), and not merely for the three validation draws.

\subsection{Smooth fitness landscape control}
In control experiments evaluating the impact of a smoothed fitness landscape, we replaced the binary success/fail validation with a graded interaction probability. The probability was scaled linearly with the average numerical distance between the program's outputs and the target values across the validation inputs:
Let \(X = \{x_1, x_2, x_3\}\) be the set of validation inputs, \(f(x)\) the target polynomial output modulo \(256\), and \(\hat{f}(x)\) the program's output. We define the normalized average circular distance \(d_{\mathrm{avg}}\) across validation draws as
\begin{align*}
    d_{\mathrm{avg}} &= \frac{1}{\lvert X \rvert} \sum_{x \in X} \frac{d_{256}(f(x), \hat{f}(x))}{128}, \\[1ex]
    \rho & = 1 - C \frac k B \\
    p(\text{interaction}) &= \max\left( p_{\mathrm{base}}, \, \left[ p_{\mathrm{succ}} - (p_{\mathrm{succ}} - p_{\mathrm{base}}) \, d_{\mathrm{avg}} \right] \cdot \rho \right),
\end{align*}
where \(d_{256}(a, b) = \min(\lvert a - b \rvert, \, 256 - \lvert a - b \rvert)\) is the circular distance modulo \(256\), \(128\) is the maximum possible circular distance, \(p_{\mathrm{succ}} = 1.0\) and \(p_{\mathrm{base}} = 0.3\) are the success and baseline interaction probabilities, and \(\rho\) is a penalty term based on \(k\), the average number of execution steps taken over the three validation runs; on \(B = 512\), the maximum instruction budget; and on \(C\), the metabolic penalty coefficient (defaulting to \(C=0.3\)).

\subsection{Simulation protocol}
Simulations run for $T_{\max} = 10^6$ epochs starting from uniformly random bytes. Parameters are summarized in Table~\ref{tab:parameters}.

\begin{table}[htbp]
\centering
\small
\caption{Simulation parameters.}
\label{tab:parameters}
\begin{tabular}{lc}
\toprule
Program length $\ell$ & 32 bytes \\
Instruction budget $B$ & 512 \\
Global grid & $512 \times 1024$ \\
Task niches $L$ & 32 (each $128 \times 128$) \\
Baseline interaction prob.\ $p_{\mathrm{base}}$ & 0.3 \\
Success interaction prob.\ $p_{\mathrm{succ}}$ & 1.0 \\
Previous-task interaction prob.\ $p_{\mathrm{prev}}$ & 0.65 \\
Metabolic penalty coeff.\ $C$ & 0.3 \\
Cross-niche pollination $\pi$ & 0.05 \\
Per-program mutation rate & 1/64 \\
Validation draws & 3 inputs, no replacement \\
Validation domain & $x \in \{0,\ldots,15\}$ \\
Output modulus & $2^8$ \\
Max epochs $T_{\max}$ & $10^6$ \\
Replicates per condition & 100 (unless otherwise stated) \\
\bottomrule
\end{tabular}
\end{table}

\SetAlFnt{\small}
\begin{algorithm}[H]
\DontPrintSemicolon
\SetKwInOut{Input}{Input}
\SetKwInOut{Output}{Output}
\Input{Grid of $512 \times 1024$ programs in $L$ niches}
\Output{Evolution trace until epoch $T_{\max}$}

Initialize all programs with random bytes\;
\For{epoch $\leftarrow 1$ \KwTo $T_{\max}$}{
  \ForEach{program $P$}{
    With probability $1/64$: mutate one random byte in $P$\;
  }
  Mark all programs as available\;
  \ForEach{program $P_1$ in random order}{
    \If{$P_1$ is available}{
      \eIf{random() $< \pi$ \emph{(CNP only)}}{
        $P_2 \leftarrow$ random program from any niche\;
      }{
        $P_2 \leftarrow$ random von Neumann neighbor of $P_1$\;
      }
      \If{$P_2$ is available and $P_2 \neq P_1$}{
        Mark $P_1$, $P_2$ unavailable\;
        Load target $f$ for $P_1$'s niche\;
        \eIf{$P_1$ produces $f(x) \bmod 2^8$ on 3 random $x \in \{0, \dots, 15\}$}{
          $p \leftarrow p_{\mathrm{succ}} - C \frac{k}{B}$\;
        }{
          $p \leftarrow p_{\mathrm{base}}$\;
        }
        With probability $p$: execute concat$(P_1, P_2)$ and split the result into two $\ell$-byte programs to replace $P_1, P_2$\;
      }
    }
  }
}
\caption{Evolution through competence-modulated interaction.}
\label{alg:main}
\end{algorithm}

To implement the pairwise interaction loop in Algorithm~\ref{alg:main} without overlapping or order bias, candidate pairs are generated each epoch by randomly shuffling all grid positions and drawing a partner for each selected program. Because programs can be drawn more than once across candidate pairings, any pair containing a duplicate program assignment is filtered out before execution to ensure that no cell is evaluated or overwritten more than once per step. Because of this filtering step, the pairing algorithm is stochastic, and a randomly varying number of programs (on average roughly 56\% of the population) are selected every step for validation and interaction.

\subsection{Replicator population dynamics with and without tasks}

To understand the extent to which imposing task validation influences the underlying methods of replication, we compared replicator population dynamics with and without tasks being applied. We present the results of this investigation in Figs.~\ref{fig:effects_of_tasks}D and \ref{fig:effects_of_tasks}E. To calculate the population counts, we used a byte-matching search to count known replicators of each type. Specifically, we scanned for the following consecutive byte sequences corresponding to key instructions for each replicator, which rely on the Z80's core block-copy registers (\texttt{HL} for source address, \texttt{DE} for destination address, and \texttt{BC} for copy length):
\begin{itemize}
    \item \textbf{LDIR} (\texttt{[0xED, 0xB0]}): The repeating block-copy instruction. Setting \(\mathtt{HL}\) to the start of parent \(P_1\), \(\mathtt{DE}\) to the start of partner \(P_2\), and \(\mathtt{BC}\) to 32 copies the entire program automatically in a single step as pointers increment and \(\mathtt{BC}\) decrements to zero.
    \item \textbf{LDDR} (\texttt{[0xED, 0xB8]}): A close relative of LDIR that decrements the pointers instead of incrementing.
    \item \textbf{LDI} (\texttt{[0xED, 0xA0]}): A non-repeating variant of LDIR that increments pointers and decrements \(\mathtt{BC}\) but does not loop automatically.
    \item \textbf{LDD} (\texttt{[0xED, 0xA8]}): A non-repeating variant that decrements pointers and decrements \(\mathtt{BC}\) but does not loop automatically, requiring an explicit external loop to achieve replication.
    \item \textbf{LoadPush families}: Rely on a sequence of paired load and stack-push instructions (e.g., loading bytes into a register and pushing them onto the stack). We tracked four variants based on the registers used:
    \begin{itemize}
        \item LoadPush (BC): \texttt{[0x01, 0xC5, 0x01, 0xC5]}
        \item LoadPush (DE): \texttt{[0x11, 0xD5, 0x11, 0xD5]}
        \item LoadPush (HL): \texttt{[0x21, 0xE5, 0x21, 0xE5]}
        \item LoadPush (HL\_2): \texttt{[0xE5, 0x2A, 0xE5, 0x2A]}
    \end{itemize}
\end{itemize}

We identified these key instructions by manually inspecting our runs and looking at execution traces to see what elements of tape were responsible for replication. 

A tape was counted if it contained the target sequence at least once. Formally, for a set of tapes \(T\) and a byte pattern \(p_v\) associated with replicator variant \(v\), the population count \(C_{v}\) is given by
\[
C_{v} = \sum_{t \in T} \mathbb{I}(p_v \subseteq t),
\]
where \(\mathbb{I}\) is the indicator function returning \(1\) if the pattern \(p_v\) exists as a contiguous subsequence in tape \(t\), and \(0\) otherwise. 

Let \(\mathcal{V}_{\mathrm{LDIR}} = \{\mathtt{LDIR}, \mathtt{LDDR}, \mathtt{LDI}, \mathtt{LDD}\}\) and \(\mathcal{V}_{\mathrm{LoadPush}} = \{\mathtt{LP_{BC}}, \mathtt{LP_{DE}}, \mathtt{LP_{HL}}, \mathtt{LP_{HL\_2}}\}\) denote the instruction variant sets for each replicator family. The overall population counts for the LDIR and LoadPush families are calculated as the sum of unique tape counts across their respective variants:
\begin{align*}
C_{\mathrm{total}}^{\mathrm{LDIR}} &= \sum_{v \in \mathcal{V}_{\mathrm{LDIR}}} C_{v}, \\[1ex]
C_{\mathrm{total}}^{\mathrm{LoadPush}} &= \sum_{v \in \mathcal{V}_{\mathrm{LoadPush}}} C_{v}.
\end{align*}

We note that this is not an exhaustive list of possible replicators, and likewise not a perfect search for replicators of this type (for instance, a NO-OP instruction could easily exist between the bytes of an LDIR, and it would functionally behave the same, but wouldn't be picked up by our search). However, in all cases the sum of all replicator families closely matches the total tape population size, so we expect the figures to be a sufficiently faithful count of the majority of replicator tapes.

To generate Fig.~\ref{fig:effects_of_tasks}D, we initialized 100 runs with different random seeds for each of the two experiment types: with task validation applied, and without it. In both settings, the number of niches and the number of tapes remained the same. For the experiments without task validation, we simply skipped the validation step (including penalizing execution length), and every tape proceeded to interaction. Note that this also raises the interaction probability of every program to 1, rather than the \(p_{\mathrm{base}}=0.3\) that non-validated programs receive under task pressure. We then computed the average population counts of the replicators across these 100 seeds every 1000 steps.

To generate Fig.~\ref{fig:effects_of_tasks}E, we repeated the same experiment as for Fig.~\ref{fig:effects_of_tasks}D, but in this case we customized the Z80 emulator we were using to turn the key instruction byte pairs for all the LDIR-family replicators except LDD into the equivalent of NO-OPs. In other words, we deactivated all LDIR-family instructions except LDD. Because the resulting transition unfolds far more slowly, these simulations were run for $10^7$ epochs. We then computed population counts the same way as for Fig.~\ref{fig:effects_of_tasks}D.

In both cases, \textit{Tasks Solved} is defined as the number of niches where $\geq 10\%$ of the tapes successfully solve the task across the entire input domain.

\subsection{Replicator robustness to mutations}
To compare how well the three replicator architectures tolerate mutation (Fig.~\ref{fig:effects_of_tasks}F), we measured whether each one retains its ability to copy itself after a number of mutation events. We seeded each test with a canonical replicator of the relevant type. The LDIR replicator was the four bytes \texttt{[0x1E, 0x20, 0xED, 0xB0]}, which set register \texttt{E} to 32 and then executed the \texttt{LDIR} instruction while relying on the default initialization \(\mathtt{HL}=0\) and \(\mathtt{BC}=0\) (which \texttt{LDIR} treats as a count of $2^{16}$, so the copy wraps harmlessly under modulo-$2\ell$ addressing until the instruction budget is exhausted), followed by 28 uniformly random bytes drawn independently in each trial to fill the tape to 32 bytes. Because the \texttt{LDIR} instruction copies all 32 bytes verbatim, the content of this trailing region does not affect replication and is reproduced faithfully along with the functional prefix. The LDD replicator was the eleven bytes \texttt{[0x2E, 0x1F, 0x1E, 0x3F, 0x0E, 0x20, 0xED, 0xA8, 0x28, 0xFC, 0x76]}, which initialize registers \texttt{L}, \texttt{E}, and \texttt{C}, run the \texttt{LDD} instruction inside a loop closed by a conditional relative jump, followed by a \texttt{HALT}, again padded with zeros to 32 bytes. The Load--Push replicator was the byte pair \texttt{[0x01, 0xC5]}, a load followed by a stack push, repeated sixteen times to fill the 32 byte tape.

Each robustness trial consisted of \(n\) successive mutation and replication cycles, with \(n \in \{1, 4, 8\}\). In every cycle we first mutated the current genome by selecting one byte position uniformly at random among its 32 positions and replacing it with a uniformly random byte in \(\{0,\ldots,255\}\), exactly as in the simulation's mutation operator. Because each position is drawn afresh, the replacement may coincide with the original byte and the same position may be selected in more than one cycle. We then placed the mutated genome in the first half of a \(2\ell\)-byte tape, set the second half to zeros, and executed the tape for up to \(B=512\) instructions under the same register initialization used during interaction. The first half of the resulting tape, which holds the program itself, became the genome carried into the next cycle. We classified a trial as a success when, on the final cycle, the second half of the executed tape, which had started as zeros, was byte-for-byte identical to the first half, indicating that the mutated program had copied all 32 of its bytes into the empty partner.

For each replicator type and each value of \(n\) we ran 100 independent trials and report the fraction that succeeded, with error bars given by 95\% Wilson score intervals. Pairwise differences between replicator types at each mutation level were assessed with one-sided two-proportion \(Z\)-tests using Bonferroni correction across the comparisons.

\subsection{Metabolic penalty sweep and halting behavior}
To probe how the metabolic constraint shapes the evolved programs (Fig.~\ref{fig:halting_behavior}), we swept the penalty coefficient \(C\) across the eight values \(0.0, 0.1, \ldots, 0.7\). For each value we ran 100 independent simulations, giving \(N=800\) runs in total. Every run used the standard configuration of 32 niches with cross-niche pollination for \(10^6\) epochs. All quantities below were measured on the final state of each grid.

For the average execution length (Fig.~\ref{fig:halting_behavior}A), we performed a single validation run for every program in the final grid. We set register \texttt{D} to one input \(x\) sampled uniformly from \(\{0,\ldots,15\}\) for each program, placed the program in the first half of a \(2\ell\)-byte tape with the second half set to zeros, and executed for up to \(B=512\) instructions, recording the number of instructions executed, which equals \(B\) when the program never halts. This per-program count is the validation execution step quantity \(k\) plotted in Fig.~\ref{fig:halting_behavior}A. We averaged this count over all programs in the grid to obtain one value per run. The 100 values at each coefficient are shown as box plots, with median, interquartile range, and whiskers at \(1.5\times\) the interquartile range, and the individual runs overlaid. The association between \(C\) and the per-run average was quantified with Spearman's rank correlation over the \(N=800\) runs.

For the halting behavior (Fig.~\ref{fig:halting_behavior}B), we restricted attention to the programs that produced the correct output on a single sampled input, that is, the programs whose validation run above returned \(\mathtt{E}=f(x) \bmod 2^8\). Each such program was executed in two environments that differ only in the value of register \texttt{D}, in both cases with the partner half of the tape set to zeros and a budget of \(B=512\) instructions. In the validation environment register \texttt{D} held the sampled input \(x\); in the interaction environment register \texttt{D} was zero, as it is during a normal interaction. In each environment we recorded whether the program reached a \texttt{HALT} within the budget or instead ran to the budget without halting, and we assigned the program to one of four mutually exclusive categories: it halted in both environments (\emph{Both}), only in the validation environment (\emph{Validation}), only in the interaction environment (\emph{Interaction}), or in neither (\emph{Neither}).

Within each run we normalized these counts to percentages of the filtered set of programs, so that the four categories sum to \(100\%\) and every run carries equal weight regardless of how many programs it contributed. Each subplot of Fig.~\ref{fig:halting_behavior}B corresponds to one category: the bar height is the median percentage across the runs at that coefficient, and the error bars are the \(95\%\) confidence interval of the median, obtained by non-parametric bootstrapping with 1000 resamples and drawing the lower and upper caps at the 2.5th and 97.5th percentiles of the resulting distribution of medians. The \emph{Validation} category corresponds to the conditional halting described in the main text, in which a program halts during validation but skips the halt during interaction. Its association with \(C\) was quantified with Spearman's rank correlation over the \(N=800\) runs.

\subsection{Computation of the ancestor contribution matrix}
% TODO: Label rule below only covers a validated P1 within its niche; specify what happens when a non-validated P1 interacts at p_base.
To trace the phylogenetic history of the emergent programs and uncover the evolutionary pathways (Fig.~\ref{fig:curriculum_dynamics}A), we tracked the niche of origin for all lineages throughout the simulation. At initialization (epoch 0), every program in the grid was labeled with an ancestor niche ID equal to its own niche $j \in \{0, \dots, L-1\}$, where $L = 32$ is the total number of niches. This label was then carried through interactions to record the most recent solved niche from which each lineage received material. When a validated program $P_1$ interacted with a neighbor $P_2$ within its own niche, the two programs written back in their place both took on $P_1$'s ancestor niche ID. When instead a cross-niche pollination event paired $P_1$ with a partner $P_2$ drawn from a different niche, the program written back in place of $P_2$ adopted the current niche index of $P_1$ as its ancestor niche ID, but only when two conditions held: $P_1$'s niche had already been marked as having solved its task, and $P_1$ had itself passed validation in that epoch. The second condition restricts the relabelling to donors that demonstrably carried a working solution, rather than to any program that happened to reside in a solved niche. A program's ancestor niche ID therefore identifies the most recent solved niche from which its active replicating lineage received material from a validated donor, defaulting to its niche of origin.

We defined a niche $i$ as having successfully solved its assigned task $i$ at the earliest epoch $t$ where the fraction of programs in the niche solving the task across the entire input domain ($16$ distinct inputs) was at least $10\%$ (corresponding to $\ge 1{,}639$ programs in a $128 \times 128$ niche). At this exact epoch, we recorded a snapshot of the ancestor niche IDs of all programs in niche $i$. Specifically, we computed a raw ancestor contribution vector for niche $i$, where the \(j\)-th component \(A_{i, j}\) represents the number of programs in niche \(i\) belonging to lineages that originated in niche \(j\):
\[
A_{i, j} = \sum_{P \in \mathcal{P}_i(t)} \mathbb{I}(\operatorname{anc}(P) = j),
\]
where \(\mathcal{P}_i(t)\) is the population of niche \(i\) at epoch \(t\) and \(\operatorname{anc}(P) \in \{0, \dots, L-1\}\) is the ancestor niche ID of program \(P\).

To aggregate this data across multiple independent runs, we performed the following normalization steps. First, for each simulation run \(g \in \{1, \dots, G\}\) (where \(G = 2000\) is the total number of random seeds), we normalized the raw counts to obtain the relative contribution of each source niche to the solution of task \(i\):
\[
N^{(g)}_{i, j} = \frac{A^{(g)}_{i, j}}{\sum_{m=0}^{L-1} A^{(g)}_{i, m}},
\]
conditioned on the task being solved in that run (\(\sum_{m=0}^{L-1} A^{(g)}_{i, m} > 0\)), and setting \(N^{(g)}_{i, j} = 0\) otherwise. Next, we computed the conditional average of these relative contributions across all runs where task \(i\) was successfully solved:
\[
V_{i, j} = \frac{\sum_{g=1}^{G} N^{(g)}_{i, j}}{\sum_{g=1}^{G} \mathbb{I}\left(\sum_{m=0}^{L-1} A^{(g)}_{i, m} > 0\right)}.
\]
The matrix \(V\) represents the average proportion of the population in niche \(i\) originating from niche \(j\) at the moment the task was solved, conditioned on task \(i\) being solved. Finally, to construct the unified heatmap (Fig.~\ref{fig:curriculum_dynamics}A), we applied a row-wise min-max normalization to highlight the relative significance of different source niches for each target task, independent of the absolute magnitude of contributions:
\[
W_{i, j} = \frac{V_{i, j} - \min_{m} V_{i, m}}{\max_{m} V_{i, m} - \min_{m} V_{i, m}},
\]
where \(W_{i, j} \in [0, 1]\) is the normalized relative contribution of source task \(j\) to target task \(i\).

\subsection{Curriculum learning}

To investigate curriculum learning, we maintain the same population structure as the CNP configuration: a population of $N_p = 524{,}288$ programs distributed across $L = 32$ separate niches (grids of $128 \times 128$ programs), with local neighborhood pairing and a global cross-niche pollination rate of $\pi = 0.05$. However, instead of each niche evaluating programs on a distinct static task from the task library, all $L$ niches in the population are evaluated on the same target task $T_i$ at any given epoch. This target task is updated dynamically as the population traverses a linear curriculum sequence \(\mathcal{C} = \langle T_1, T_2, \dots, T_M \rangle\), which is defined as either \(\mathcal{C}_{\text{optimal}} = \langle 2n, \, n^2+n, \, 2n^2+n, \, n^3+n^2+n, \, n^3+n^2+n+3 \rangle\), or \(\mathcal{C}_{\text{simple}} = \langle n, \, n^2, \, n^3, \, n^3+n, \, n^3+n^2+n, \, n^3+n^2+n+3 \rangle\).

To enable evolutionary scaffolding and prevent the loss of capabilities acquired in the previous step, the competence-modulated interaction probability is defined as
\begin{equation*}
    p(\text{interaction}) = 
    \begin{cases}
    p_{\mathrm{succ}} - C \dfrac{k}{B} & \text{if validated on active task } T_i, \\[6pt]
    p_{\mathrm{prev}} - C \dfrac{k}{B} & \text{if validated on previous task } T_{i-1}, \\[6pt]
    p_{\mathrm{base}} & \text{otherwise},
    \end{cases}
\end{equation*}
where \(T_i\) is the active task, \(T_{i-1}\) is the immediately preceding task in the curriculum (\(T_0\) defaulting to not validated), \(p_{\mathrm{succ}} = 1.0\) is the success interaction probability, \(p_{\mathrm{prev}} = 0.65\) is the interaction probability for succeeding on the previous task, \(p_{\mathrm{base}} = 0.3\) is the baseline interaction probability, \(k\) is the average number of execution steps taken over the three validation runs, \(B = 512\) is the maximum instruction budget, and \(C\) is the metabolic penalty coefficient.

% TODO: The curriculum advances when a single program solves the task, whereas niches count as solved at 10%. State how "success rate per task" in Fig. 5B is defined for the curriculum runs.
The active curriculum task advances globally from $T_i$ to $T_{i+1}$ (setting the target task for all niches to $T_{i+1}$) at the next logging checkpoint (performed every $1{,}000$ epochs) if at least one program in any of the niches successfully solves the active task in its complete domain. The total computational budget is kept identical to the CNP benchmark runs.

\section*{Author Contributions}
F.C., E.N., E.R., and B.A.R. designed research; F.C., E.N., E.R., and S.B. performed research; F.C., E.N., and A.B. analyzed data; and F.C., E.N., E.R., S.B., A.B., M.E., R.S., B.L., J.M., B.A.A., and B.A.R. wrote the paper.

\bibliography{bibliography}

\clearpage
\appendix

\begin{center}
   \LARGE\textbf{Supplementary Information}
\end{center}
\vspace{1.5em}

% Define color palette for annotated Z80 tape disassembly
\definecolor{zrep}{HTML}{1F77B4}    % Steel blue: replication preamble
\definecolor{zinit}{HTML}{2CA02C}   % Forest green: loop & register initialization
\definecolor{zarith}{HTML}{D95F02}  % Rust / amber: polynomial arithmetic & loops
\definecolor{zhalt}{HTML}{7570B3}   % Slate purple: output formatting & halt
\definecolor{zunused}{HTML}{888888} % Muted gray: unused trailing bytes

\section{Task performance comparison: emergent self-replication vs. hard-wired copy}\label{app:control_performance}

In Sec.~\ref{subsec:curriculum_structure}, we examined how emergent self-replication influences the ancestral curriculum structure compared to a control setup with hard-wired reproduction. In the hard-wired copy control, pairwise interactions ($P_1 + P_2 \to P_1 + P_1$) overwrite the partner program directly with an exact copy of the first without executing Z80 code, and memory writes during validation are discarded. This leaves the entire 32-byte tape available for computation and eliminates replication errors.

\begin{figure}[htbp]
   \centering
   \includegraphics[width=\linewidth]{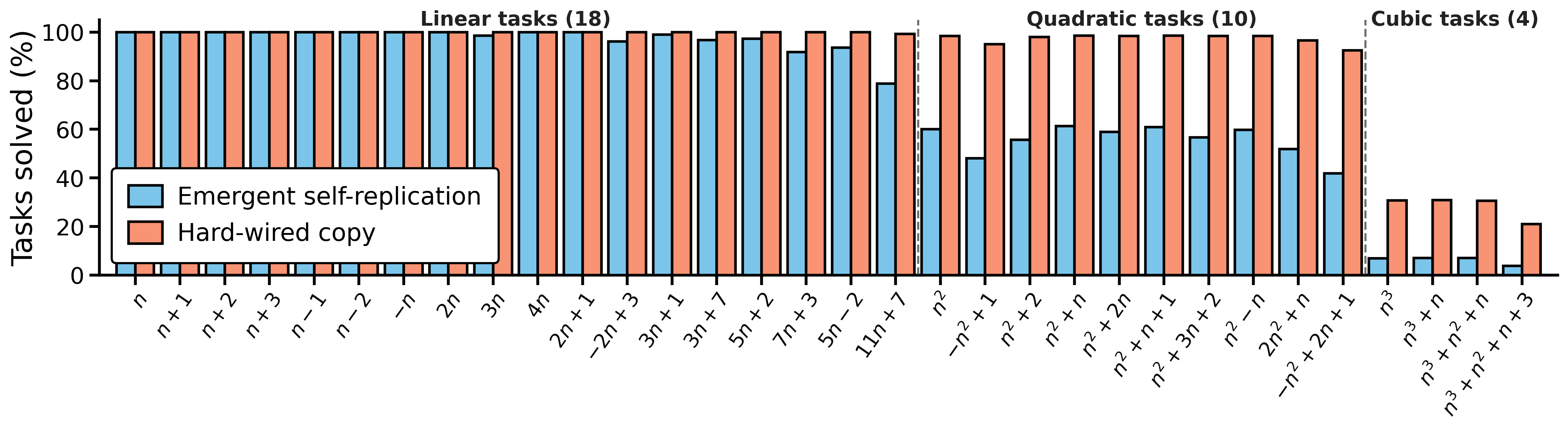}
   \caption{\textbf{Task solve percentages across 32 polynomial tasks for emergent self-replication versus hard-wired copy.} Percentage of independent runs ($N = 2{,}000$ for emergent self-replication, $N = 2{,}000$ for hard-wired copy) that solve each polynomial task after $10^6$ epochs. Dashed vertical lines indicate the division between linear (tasks 0--17), quadratic (tasks 18--27), and cubic (tasks 28--31) polynomials. The hard-wired copy control solves a higher percentage of quadratic and cubic tasks due to faithful copying and dedicated tape capacity.}
   \label{fig:supp_performance_comparison}
\end{figure}

Fig.~\ref{fig:supp_performance_comparison} shows the percentage of independent runs that solved each of the 32 polynomial tasks under emergent self-replication ($N = 2{,}000$) versus hard-wired copy ($N = 2{,}000$) after $10^6$ epochs.

While both systems solve the 18 linear tasks (tasks 0--17) in the large majority of runs, the hard-wired copy system achieves higher solve rates on quadratic tasks (tasks 18--27, averaging $>96\%$ solve rate vs. $\sim 50\text{--}60\%$ in the emergent system) and cubic tasks (tasks 28--31, achieving $21.0\text{--}30.8\%$ solve rate vs. $3.8\text{--}7.0\%$ in the emergent system). This performance difference confirms that the wider genealogical breadth observed under emergent self-replication (Fig.~\ref{fig:curriculum_dynamics}C, D) is not driven by higher task-solving competence, but reflects the broader set of evolutionary routes opened when replication mechanisms themselves vary and coevolve alongside task logic.

\section{Evolved program architectures and Z80 assemblies}\label{app:discovered_architectures}

The coevolution of self-replication and task evaluation produces distinct program structures in the two setups. Below we examine representative 32-byte Z80 programs evolved for linear, quadratic, and cubic tasks, including pure monomials ($n^2$, $n^3$) and multi-term polynomials ($n+1$, $2n$, $n^2+n$, $n^3+n^2+n$). Each example includes the complete 32-byte hexadecimal tape alongside its annotated disassembly (Secs.~\ref{subsubsec:task1}--\ref{subsubsec:task30}).

\paragraph{Emergent self-replication architecture.}
In the default system, programs must share their 32-byte tape between self-replication logic and the arithmetic operations needed for polynomial evaluation. Evolved programs consistently develop a compact block-copy preamble centered on the Z80's \texttt{LDIR} instruction (\texttt{ED B0}). By setting up pointer and counter registers (e.g., \texttt{LD C, 2Eh; LDIR}), the program copies its 32-byte sequence into the partner cell during interaction (counter values exceeding 32 overrun harmlessly under mod-64 addressing, rewriting already-copied bytes with identical values).

Because the Z80 lacks a dedicated multiply instruction, polynomial arithmetic is implemented through register addition and shift loops:
\begin{itemize}
   \item For linear tasks (e.g., $f(n)=2n$, Sec.~\ref{subsubsec:task7}), programs double the input via 16-bit addition (\texttt{ADD HL, HL}) after loading register \texttt{D} into \texttt{HL}, placing the result into register \texttt{E} with \texttt{EX DE, HL} or \texttt{LD E, L}.
% TODO: The instruction sequence quoted here does not obviously yield 7n+3; give the actual listing or drop the example.
   \item For scaling tasks (e.g., $f(n)=7n+3$), programs combine additions across \texttt{HL} and \texttt{BC} (\texttt{ADD HL, BC; ADD HL, HL; ADD HL, BC}) to achieve multi-step linear scaling in few instructions.
   \item For quadratic tasks ($n^2$, $n^2+n$, Secs.~\ref{subsubsec:task18} and \ref{subsubsec:task21}), programs initialize an accumulation loop using the hardware decrement-and-jump instruction \texttt{DJNZ} (\texttt{0x10}), repeatedly adding $n$ to an accumulator.
   \item For cubic tasks ($n^3$, $n^3+n^2+n$, Secs.~\ref{subsubsec:task28} and \ref{subsubsec:task30}), programs maintain nested or chained accumulation loops across registers \texttt{B}, \texttt{C}, \texttt{D}, and \texttt{A}.
\end{itemize}
Programs terminate execution with a \texttt{HALT} instruction (\texttt{0x76}) once the result is placed in register \texttt{E}, minimizing execution steps during validation under the metabolic pressure described in Sec.~\ref{subsec:metabolism}.

\paragraph{Hard-wired copy architecture.}
In the control system, replication is handled by the environment, so programs omit the \texttt{LDIR} preamble entirely and use the full 32 bytes for computation:
\begin{itemize}
   \item For simple linear tasks (e.g., $f(n)=n+1$, Sec.~\ref{subsubsec:task1}), solutions require only 3 bytes: loading register \texttt{D} into \texttt{E} (\texttt{LD E, D}), incrementing (\texttt{INC E}), and halting (\texttt{HALT}) in 3 CPU steps.
   \item For pure quadratic tasks such as $f(n)=n^2$ (Sec.~\ref{subsubsec:task18}), programs accumulate $n$ for $n+1$ iterations (\texttt{ADD A, D; DJNZ loop}) yielding $n(n+1)$, followed by a single subtraction (\texttt{SUB D}) to reach $n^2$ in 9 bytes.
   \item For compound quadratic tasks such as $f(n)=n^2+n=n(n+1)$ (Sec.~\ref{subsubsec:task21}), programs initialize register \texttt{B} with $n+1$ (\texttt{LD B, D; INC B}) and accumulate $n$ directly, computing the polynomial in 8 bytes (\texttt{8D 42 04 82 10 BD 5F 76}).
   \item For cubic tasks ($n^3$, $n^3+n^2+n$, Secs.~\ref{subsubsec:task28} and \ref{subsubsec:task30}), programs chain successive \texttt{DJNZ} loops and stack-dispatch restarts, using intermediate register values (\texttt{LD B, E; LD E, A}) to compute higher-order products.
\end{itemize}

\paragraph{Validation cost metric.}
Validation cost denotes the number of CPU instruction cycles/steps executed until \texttt{HALT} is reached. During evolutionary evaluation, validation samples three inputs from $x \in \{0, 1, \dots, 15\}$ (Sec.~\ref{subsec:methods_competence}); the costs reported here are instead computed across all 16 test inputs. In linear programs with straight-line execution, the step count is input-invariant; in quadratic and cubic programs containing data-dependent loops (e.g., \texttt{DJNZ} looping $n$ or $n+1$ times), the reported validation cost is the mean execution steps averaged over all 16 test inputs.

\paragraph{Conditional replication skipping during validation.}
In some lineages, programs evolve to distinguish between the validation and interaction phases by exploiting the initialization of register \texttt{D}. In our validation protocol (Sec.~\ref{subsec:methods_competence}), a candidate program $P_1$ is loaded into the first 32 bytes of memory while the second 32-byte half is initialized to zeros. While modifications $P_1$ makes to its own 32-byte space during validation persist, the trailing 32-byte partner buffer is discarded at the end of the run. Executing the 32-byte \texttt{LDIR} block copy during validation therefore copies data into a temporary buffer with no persisting reproductive effect, while consuming execution steps and incurring a metabolic penalty. Because register \texttt{D} holds the test input $x \in \{0, \dots, 15\}$ during validation (which is non-zero in 15 out of 16 trials) but is initialized to zero ($D = 0$) during interaction, evolving lineages can develop conditional control flow on \texttt{D} that avoids executing the copy during validation. In the representative task-18 and task-21 programs (Secs.~\ref{subsubsec:task18} and \ref{subsubsec:task21}), this gating operates through overlapping instruction decoding: on the fall-through path, the replication bytes \texttt{ED B0} are consumed as the address operand of a 4-byte \texttt{LD BC, (nnnn)} instruction, so no copy is performed. The program then adds \texttt{D} into the accumulator; when $D = 0$ (the interaction phase), a conditional relative jump re-enters the same byte region at a shifted offset, where the bytes decode as \texttt{LD C, E; LDIR} and execute the full replication. This phase distinction is related to the conditional halting dynamics discussed in Sec.~\ref{subsec:metabolism}. By saving the 32--46 CPU steps of the copy during validation, such gating explains why some quadratic solvers execute in fewer total steps than linear programs that happen to retain unconditional block copies (such as the task 7 example in Sec.~\ref{subsubsec:task7}).

\subsection{Representative program listings}\label{subsec:representative_listings}

The listings below compare complete 32-byte programs evolved under emergent self-replication and hard-wired copy across task complexities. Annotations assume the register initialization of Sec.~\ref{subsec:methods_competence}, with the flags register initialized (like \texttt{A}) to \texttt{FFh}, so the zero and carry flags are set at entry. All jump targets, memory operands, and stack accesses resolve modulo the 64-byte address space; offsets \texttt{20h}--\texttt{3Fh} denote the second (partner) half of memory, which holds a clone of the tape after a successful \texttt{LDIR}. Instruction boundaries follow the executed path: bytes skipped by dispatch or consumed as operands are marked accordingly. Functional sections are color-coded consistently across the raw hexadecimal tape and the annotated disassembly:

\begin{center}
\footnotesize
\textbf{Functional color coding:}\quad
\textcolor{zrep}{\textbf{\textbullet\ Replication preamble}}\quad
\textcolor{zinit}{\textbf{\textbullet\ Register/loop setup}}\quad
\textcolor{zarith}{\textbf{\textbullet\ Arithmetic/loop body}}\quad
\textcolor{zhalt}{\textbf{\textbullet\ Output \& halt}}\quad
\textcolor{zunused}{\textbf{\textbullet\ Unexecuted / trailing bytes}}
\end{center}

% ----------------------------------------------------------------------
% Task 1: Linear f(n) = n + 1
% ----------------------------------------------------------------------
\subsubsection{Linear task: \texorpdfstring{$f(n)=n+1$}{f(n) = n+1} (task 1)}\label{subsubsec:task1}

\noindent\begin{minipage}[t]{0.48\textwidth}
\textbf{Emergent self-replication} \par\vspace{0.3em}
\textbf{Hexadecimal tape (32 bytes):} \par
{\scriptsize\ttfamily\raggedright
\textcolor{zrep}{E0 5E 0E 09 ED B0} \textcolor{zarith}{14} \textcolor{zhalt}{5A 76} \textcolor{zunused}{41 00 41 00 41 00 41} \par
\textcolor{zunused}{00 41 00 41 00 41 00 41 00 41 00 41 00 41 00 41} \par
} \vspace{0.4em}
\textbf{Disassembly:} \par
{\footnotesize\ttfamily\raggedright
\textcolor{zrep}{00: E0       RET PO       ; not taken (P/V = 1 at entry)} \par
\textcolor{zrep}{01: 5E       LD E, (HL)   ; E = mem[00h] = E0h == 20h mod 64 (partner offset)} \par
\textcolor{zrep}{02: 0E 09    LD C, 09h    ; BC = 9 (byte counter)} \par
\textcolor{zrep}{04: ED B0    LDIR         ; copy the 9-byte functional core to the partner half} \par
\textcolor{zarith}{06: 14       INC D        ; D = n + 1} \par
\textcolor{zhalt}{07: 5A       LD E, D      ; E = n + 1 (output register)} \par
\textcolor{zhalt}{08: 76       HALT         ; halt execution} \par
\textcolor{zunused}{09--1F:      [Trailing padding: 41 00 \dots]} \par
}
\end{minipage}
\hfill
\begin{minipage}[t]{0.48\textwidth}
\textbf{Hard-wired copy} \par\vspace{0.3em}
\textbf{Hexadecimal tape (32 bytes):} \par
{\scriptsize\ttfamily\raggedright
\textcolor{zinit}{5A} \textcolor{zarith}{1C} \textcolor{zhalt}{76} \textcolor{zunused}{36 C5 A3 1B 94 82 A2 49 4D B8 32 AF 18} \par
\textcolor{zunused}{4E FB 3B 31 8B E3 26 AA 66 78 A3 93 99 05 37 58} \par
} \vspace{0.4em}
\textbf{Disassembly:} \par
{\footnotesize\ttfamily\raggedright
\textcolor{zinit}{00: 5A       LD E, D      ; E = input n} \par
\textcolor{zarith}{01: 1C       INC E        ; E = n + 1} \par
\textcolor{zhalt}{02: 76       HALT         ; halt execution} \par
\textcolor{zunused}{03--1F:      [Unused trailing genomic bytes]} \par
}
\end{minipage}

\vspace{0.4em}
\noindent\begin{minipage}[t]{0.48\textwidth}
\textit{\small Validation cost: 15 CPU steps (constant for all $x$).}
\end{minipage}\hfill
\begin{minipage}[t]{0.48\textwidth}
\textit{\small Validation cost: 3 CPU steps (constant for all $x$).}
\end{minipage}

\vspace{1.2em}

% ----------------------------------------------------------------------
% Task 7: Linear f(n) = 2n
% ----------------------------------------------------------------------
\subsubsection{Linear scaling task: \texorpdfstring{$f(n)=2n$}{f(n) = 2n} (task 7)}\label{subsubsec:task7}

\noindent\begin{minipage}[t]{0.48\textwidth}
\textbf{Emergent self-replication} \par\vspace{0.3em}
\textbf{Hexadecimal tape (32 bytes):} \par
{\scriptsize\ttfamily\raggedright
\textcolor{zrep}{A0 5E 0E 2E ED B0} \textcolor{zinit}{09 94 6A} \textcolor{zarith}{29} \textcolor{zhalt}{EB 76} \textcolor{zunused}{0C 0D 56 47} \par
\textcolor{zunused}{3D 31 D3 46 8A 08 BD 5F 58 D4 2C 19 E8 CD FF 4F} \par
} \vspace{0.4em}
\textbf{Disassembly:} \par
{\footnotesize\ttfamily\raggedright
\textcolor{zrep}{00: A0       AND B        ; A = 0 (B = 0); clears carry} \par
\textcolor{zrep}{01: 5E       LD E, (HL)   ; E = mem[00h] = A0h == 20h mod 64 (partner offset)} \par
\textcolor{zrep}{02: 0E 2E    LD C, 2Eh    ; BC = 46 (byte counter)} \par
\textcolor{zrep}{04: ED B0    LDIR         ; copy tape to partner (46-byte count overruns harmlessly mod 64)} \par
\textcolor{zinit}{06: 09       ADD HL, BC   ; no-op (BC = 0 after LDIR)} \par
\textcolor{zinit}{07: 94       SUB H        ; A = 0 (accumulator clear)} \par
\textcolor{zinit}{08: 6A       LD L, D      ; HL = input n (H = 0)} \par
\textcolor{zarith}{09: 29       ADD HL, HL   ; HL = 2n (16-bit shift)} \par
\textcolor{zhalt}{0A: EB       EX DE, HL    ; DE = 2n; result in register E} \par
\textcolor{zhalt}{0B: 76       HALT         ; halt execution} \par
\textcolor{zunused}{0C--1F:      [Unused trailing genomic bytes]} \par
}
\end{minipage}
\hfill
\begin{minipage}[t]{0.48\textwidth}
\textbf{Hard-wired copy} \par\vspace{0.3em}
\textbf{Hexadecimal tape (32 bytes):} \par
{\scriptsize\ttfamily\raggedright
\textcolor{zinit}{F6 44 7A} \textcolor{zarith}{87} \textcolor{zhalt}{5F 76} \textcolor{zunused}{F8 1F 8E 47 C1 48 74 CD 3D 1A} \par
\textcolor{zunused}{C3 5D 55 60 20 20 11 DB C1 74 A0 FA BE D2 05 71} \par
} \vspace{0.4em}
\textbf{Disassembly:} \par
{\footnotesize\ttfamily\raggedright
\textcolor{zinit}{00: F6 44    OR 44h       ; no functional effect (A overwritten below)} \par
\textcolor{zinit}{02: 7A       LD A, D      ; A = input n} \par
\textcolor{zarith}{03: 87       ADD A, A     ; A = 2n (doubling)} \par
\textcolor{zhalt}{04: 5F       LD E, A      ; E = 2n (output register)} \par
\textcolor{zhalt}{05: 76       HALT         ; halt execution} \par
\textcolor{zunused}{06--1F:      [Unused trailing genomic bytes]} \par
}
\end{minipage}

\vspace{0.4em}
\noindent\begin{minipage}[t]{0.48\textwidth}
\textit{\small Validation cost: 55 CPU steps (46 replication steps + 9 arithmetic steps; constant for all $x$).}
\end{minipage}\hfill
\begin{minipage}[t]{0.48\textwidth}
\textit{\small Validation cost: 5 CPU steps (constant for all $x$).}
\end{minipage}

\vspace{1.2em}

% ----------------------------------------------------------------------
% Task 18: Pure Quadratic f(n) = n^2
% ----------------------------------------------------------------------
\subsubsection{Pure quadratic task: \texorpdfstring{$f(n)=n^2$}{f(n) = n\textasciicircum 2} (task 18)}\label{subsubsec:task18}

\noindent\begin{minipage}[t]{0.48\textwidth}
\textbf{Emergent self-replication} \par\vspace{0.3em}
\textbf{Hexadecimal tape (32 bytes):} \par
{\scriptsize\ttfamily\raggedright
\textcolor{zrep}{20 F2 5E ED 4B ED B0} \textcolor{zinit}{8A 28 7A D7} \textcolor{zunused}{CC 27 BF D1 10} \par
\textcolor{zinit}{47 A5 20 90} \textcolor{zarith}{82 10 BD} \textcolor{zhalt}{5F 76} \textcolor{zunused}{89 98 32 FF 0B 00 79} \par
} \vspace{0.4em}
\textbf{Disassembly:} \par
{\footnotesize\ttfamily\raggedright
\textcolor{zrep}{00: 20 F2    JR NZ, 34h   ; entry branch (not taken; ZF = 1 at entry)} \par
\textcolor{zrep}{02: 5E       LD E, (HL)   ; E = mem[00h] = 20h (partner offset)} \par
\textcolor{zrep}{03: ED 4B ED B0} \par
\textcolor{zrep}{\phantom{03:} LD BC, (B0EDh) ; 4-byte load consuming ED B0 as its address operand -- no LDIR executes on this path; BC = mem[2Dh..2Eh] (= 0 during validation)} \par
\textcolor{zinit}{07: 8A       ADC A, D     ; A = n (FFh + n + CF)} \par
\textcolor{zinit}{08: 28 7A    JR Z, 04h    ; taken when D = 0 (interaction): re-enters at 04h, where 4B; ED B0 decode as LD C, E; LDIR -- the replication path} \par
\textcolor{zinit}{0A: D7       RST 10h      ; dispatch: jump to 10h (return address pushed into the partner half; bytes 0Bh--0Fh never execute)} \par
\textcolor{zinit}{10: 47       LD B, A      ; B = n (loop counter)} \par
\textcolor{zinit}{11: A5       AND L        ; A = 0 (L = 0); clears carry} \par
\textcolor{zinit}{12: 20 90    JR NZ, 24h   ; not taken (ZF = 1)} \par
\textcolor{zarith}{14: 82       ADD A, D     ; loop body: A = A + n} \par
\textcolor{zarith}{15: 10 BD    DJNZ 14h     ; loop n times: A = n\string^2} \par
\textcolor{zhalt}{17: 5F       LD E, A      ; E = n\string^2 (output register)} \par
\textcolor{zhalt}{18: 76       HALT         ; halt execution} \par
\textcolor{zunused}{19--1F:      [Unused trailing bytes: 89 98 32 \dots]} \par
}
\end{minipage}
\hfill
\begin{minipage}[t]{0.48\textwidth}
\textbf{Hard-wired copy} \par\vspace{0.3em}
\textbf{Hexadecimal tape (32 bytes):} \par
{\scriptsize\ttfamily\raggedright
\textcolor{zinit}{7D 42 04} \textcolor{zarith}{82 10 FD 92} \textcolor{zhalt}{5F 76} \textcolor{zunused}{18 03 DC 9B 4B CB 4C} \par
\textcolor{zunused}{31 74 A9 74 E2 A1 B5 71 1F 1D 61 3F 79 9D D2 28} \par
} \vspace{0.4em}
\textbf{Disassembly:} \par
{\footnotesize\ttfamily\raggedright
\textcolor{zinit}{00: 7D       LD A, L      ; A = 0 (accumulator clear)} \par
\textcolor{zinit}{01: 42       LD B, D      ; B = input n} \par
\textcolor{zinit}{02: 04       INC B        ; B = n + 1 (loop counter)} \par
\textcolor{zarith}{03: 82       ADD A, D     ; loop target: A = A + n} \par
\textcolor{zarith}{04: 10 FD    DJNZ 03h     ; loop n+1 times (A = n(n+1))} \par
\textcolor{zarith}{06: 92       SUB D        ; A = n(n+1) - n = n\string^2} \par
\textcolor{zhalt}{07: 5F       LD E, A      ; E = n\string^2 (output register)} \par
\textcolor{zhalt}{08: 76       HALT         ; halt execution} \par
\textcolor{zunused}{09--1F:      [Unused trailing genomic bytes]} \par
}
\end{minipage}

\vspace{0.4em}
\noindent\begin{minipage}[t]{0.48\textwidth}
\textit{\small Validation cost: $\sim$26 CPU steps (mean over $x \in \{0, \dots, 15\}$).}
\end{minipage}\hfill
\begin{minipage}[t]{0.48\textwidth}
\textit{\small Validation cost: 23 CPU steps (mean over $x \in \{0, \dots, 15\}$).}
\end{minipage}

\vspace{1.2em}

% ----------------------------------------------------------------------
% Task 21: Quadratic f(n) = n^2 + n
% ----------------------------------------------------------------------
\subsubsection{Quadratic task: \texorpdfstring{$f(n)=n^2+n$}{f(n) = n\textasciicircum 2+n} (task 21)}\label{subsubsec:task21}

\noindent\begin{minipage}[t]{0.48\textwidth}
\textbf{Emergent self-replication} \par\vspace{0.3em}
\textbf{Hexadecimal tape (32 bytes):} \par
{\scriptsize\ttfamily\raggedright
\textcolor{zrep}{20 FA 5E ED 4B ED B0} \textcolor{zinit}{8A 28 FA 1E D8 01 93 A5 47} \par
\textcolor{zinit}{3D EA 2C 08} \textcolor{zarith}{8A 10 BD} \textcolor{zhalt}{5F 76} \textcolor{zunused}{46 4F 12 48 17 47 68} \par
} \vspace{0.4em}
\textbf{Disassembly:} \par
{\footnotesize\ttfamily\raggedright
\textcolor{zrep}{00: 20 FA    JR NZ, 3Ch   ; entry branch (not taken; ZF = 1 at entry)} \par
\textcolor{zrep}{02: 5E       LD E, (HL)   ; E = mem[00h] = 20h (partner offset)} \par
\textcolor{zrep}{03: ED 4B ED B0} \par
\textcolor{zrep}{\phantom{03:} LD BC, (B0EDh) ; 4-byte load consuming ED B0 as its address operand -- no LDIR on this path; BC = mem[2Dh..2Eh]} \par
\textcolor{zinit}{07: 8A       ADC A, D     ; A = n (FFh + n + CF); leaves CF = 1} \par
\textcolor{zinit}{08: 28 FA    JR Z, 04h    ; taken when D = 0 (interaction): re-enters at 04h as LD C, E; LDIR -- the replication path} \par
\textcolor{zinit}{0A: 1E D8    LD E, D8h    ; scratch load (overwritten below)} \par
\textcolor{zinit}{0C: 01 93 A5 LD BC, A593h ; register-pair setup (B overwritten below)} \par
\textcolor{zinit}{0F: 47       LD B, A      ; B = n (loop counter)} \par
\textcolor{zinit}{10: 3D       DEC A        ; A = n - 1 (CF unaffected)} \par
\textcolor{zinit}{11: EA 2C 08 JP PE, 082Ch ; not taken (no overflow for n <= 15)} \par
\textcolor{zarith}{14: 8A       ADC A, D     ; loop body: A = A + n (+CF on the first pass)} \par
\textcolor{zarith}{15: 10 BD    DJNZ 14h     ; loop n times: A = (n-1) + (n+1) + (n-1)n = n\string^2 + n} \par
\textcolor{zhalt}{17: 5F       LD E, A      ; E = n\string^2 + n} \par
\textcolor{zhalt}{18: 76       HALT         ; halt execution} \par
\textcolor{zunused}{19--1F:      [Unused trailing bytes: 46 4F 12 \dots]} \par
}
\end{minipage}
\hfill
\begin{minipage}[t]{0.48\textwidth}
\textbf{Hard-wired copy} \par\vspace{0.3em}
\textbf{Hexadecimal tape (32 bytes):} \par
{\scriptsize\ttfamily\raggedright
\textcolor{zinit}{8D 42 04} \textcolor{zarith}{82 10 BD} \textcolor{zhalt}{5F 76} \textcolor{zunused}{7A F9 10 D9 EE 16 4E 32} \par
\textcolor{zunused}{B4 B8 4A A8 A5 58 D3 51 52 B4 09 6E D8 B7 9B 03} \par
} \vspace{0.4em}
\textbf{Disassembly:} \par
{\footnotesize\ttfamily\raggedright
\textcolor{zinit}{00: 8D       ADC A, L     ; A = 0 (FFh + 0 + CF)} \par
\textcolor{zinit}{01: 42       LD B, D      ; B = input n} \par
\textcolor{zinit}{02: 04       INC B        ; B = n + 1 (loop counter)} \par
\textcolor{zarith}{03: 82       ADD A, D     ; loop target: A = A + n} \par
\textcolor{zarith}{04: 10 BD    DJNZ 03h     ; loop n+1 times: A = n(n+1)} \par
\textcolor{zhalt}{06: 5F       LD E, A      ; E = n\string^2 + n} \par
\textcolor{zhalt}{07: 76       HALT         ; halt execution} \par
\textcolor{zunused}{08--1F:      [Unused trailing genomic bytes]} \par
}
\end{minipage}

\vspace{0.4em}
\noindent\begin{minipage}[t]{0.48\textwidth}
\textit{\small Validation cost: $\sim$25 CPU steps (mean over $x \in \{0, \dots, 15\}$).}
\end{minipage}\hfill
\begin{minipage}[t]{0.48\textwidth}
\textit{\small Validation cost: 22 CPU steps (mean over $x \in \{0, \dots, 15\}$).}
\end{minipage}

\vspace{1.2em}

% ----------------------------------------------------------------------
% Task 28: Pure Cubic f(n) = n^3
% ----------------------------------------------------------------------
\subsubsection{Pure cubic task: \texorpdfstring{$f(n)=n^3$}{f(n) = n\textasciicircum 3} (task 28)}\label{subsubsec:task28}

\noindent\begin{minipage}[t]{0.48\textwidth}
\textbf{Emergent self-replication} \par\vspace{0.3em}
\textbf{Hexadecimal tape (32 bytes):} \par
{\scriptsize\ttfamily\raggedright
\textcolor{zrep}{20 36 5E 7A 4B ED B0} \textcolor{zinit}{84 B0 19 5F 1C 43 FC 1C F3} \par
\textcolor{zinit}{57 CC 9C CE} \textcolor{zarith}{92 10 1D 43 5F} \textcolor{zhalt}{C0 76} \textcolor{zunused}{47 89 06 FB B0} \par
} \vspace{0.4em}
\textbf{Disassembly:} \par
{\footnotesize\ttfamily\raggedright
\textcolor{zrep}{00: 20 36    JR NZ, 38h   ; entry branch (not taken; ZF = 1 at entry)} \par
\textcolor{zrep}{02: 5E       LD E, (HL)   ; E = mem[00h] = 20h (partner offset)} \par
\textcolor{zrep}{03: 7A       LD A, D      ; A = input n} \par
\textcolor{zrep}{04: 4B       LD C, E      ; BC = 32 (byte counter)} \par
\textcolor{zrep}{05: ED B0    LDIR         ; copy tape to partner half (20h--3Fh)} \par
\textcolor{zinit}{07: 84       ADD A, H     ; A = n (H = 0); clears carry} \par
\textcolor{zinit}{08: B0       OR B         ; A = n (B = 0)} \par
\textcolor{zinit}{09: 19       ADD HL, DE   ; pointer adjustment (no functional role)} \par
\textcolor{zinit}{0A: 5F       LD E, A      ; E = n} \par
\textcolor{zinit}{0B: 1C       INC E        ; E = n + 1} \par
\textcolor{zinit}{0C: 43       LD B, E      ; B = n + 1 (loop counter)} \par
\textcolor{zinit}{0D: FC 1C F3 CALL M, F31Ch; not taken (S = 0)} \par
\textcolor{zinit}{10: 57       LD D, A      ; D = accumulator at pass entry (pass 1: D = n)} \par
% TODO: Z is set after OR B when n = 0, so this CALL is taken for n = 0 (target 1Ch mod 64); confirm the n = 0 path.
\textcolor{zinit}{11: CC 9C CE CALL Z, CE9Ch; not taken} \par
\textcolor{zarith}{14: 92       SUB D        ; loop body: A = A - D} \par
\textcolor{zarith}{15: 10 1D    DJNZ 34h     ; target 34h = 14h + 20h: the loop alternates between this SUB/DJNZ pair and its identical clone in the partner half; n+1 iterations give A = A - (n+1)D} \par
\textcolor{zarith}{17: 43       LD B, E      ; reload counter (B = n + 1)} \par
\textcolor{zarith}{18: 5F       LD E, A      ; E = accumulator (pass 1: -n\string^2; pass 2: n\string^3)} \par
\textcolor{zhalt}{19: C0       RET NZ       ; return-oriented dispatch: pops genome words as return addresses, re-entering at 30h for a second pass with D = -n\string^2 (yielding A = n\string^3), then dispatching to the cloned HALT at 3Ah} \par
\textcolor{zhalt}{1A: 76       HALT         ; E = n\string^3} \par
\textcolor{zunused}{1B--1F:      [Unused trailing bytes: 47 89 06 \dots]} \par
}
\end{minipage}
\hfill
\begin{minipage}[t]{0.48\textwidth}
\textbf{Hard-wired copy} \par\vspace{0.3em}
\textbf{Hexadecimal tape (32 bytes):} \par
{\scriptsize\ttfamily\raggedright
\textcolor{zinit}{B1 4B 04} \textcolor{zarith}{82 10 3D FD 43 5F 52 3A 24 22 BC 08 D0} \par
\textcolor{zarith}{E2 18} \textcolor{zhalt}{76} \textcolor{zunused}{12 24 24 BB 64 43 51 F8 68 CB 6D 27 83} \par
} \vspace{0.4em}
\textbf{Disassembly:} \par
{\footnotesize\ttfamily\raggedright
\textcolor{zinit}{00: B1       OR C         ; A = FFh (C = 0); clears carry} \par
\textcolor{zinit}{01: 4B       LD C, E      ; C = E} \par
\textcolor{zinit}{02: 04       INC B        ; B = B + 1 (round counter)} \par
\textcolor{zarith}{03: 82       ADD A, D     ; accumulation step: A = A + n} \par
\textcolor{zarith}{04: 10 3D    DJNZ 03h     ; loop control (target 43h == 03h mod 64)} \par
\textcolor{zarith}{06: FD 43    LD B, E      ; FD-prefixed (prefix without effect): B = E} \par
\textcolor{zarith}{08: 5F       LD E, A      ; E = accumulator} \par
\textcolor{zarith}{09: 52       LD D, D      ; no-op} \par
\textcolor{zarith}{0A: 3A 24 22 LD A, (2224h); A = mem[24h mod 64] (scratch half)} \par
\textcolor{zarith}{0D: BC       CP H         ; flag setup} \par
\textcolor{zarith}{0E: 08       EX AF, AF'   ; swap accumulator/flag set with shadow} \par
\textcolor{zarith}{0F: D0       RET NC       ; stack dispatch: pops mem[3Fh], mem[00h] -> PC = 00h, restarting the round with updated registers} \par
\textcolor{zarith}{10: E2 18 76 JP PO, 7618h ; conditional dispatch (target == 18h mod 64); the operand byte 76h at offset 12h doubles as a HALT when execution is dispatched there} \par
\textcolor{zunused}{13--1F:      [Unused trailing genomic bytes]} \par
}
\end{minipage}

\vspace{0.4em}
\noindent\begin{minipage}[t]{0.48\textwidth}
\textit{\small Validation cost: $\sim$92 CPU steps (mean over $x \in \{0, \dots, 15\}$).}
\end{minipage}\hfill
\begin{minipage}[t]{0.48\textwidth}
\textit{\small Validation cost: 48 CPU steps (mean over $x \in \{0, \dots, 15\}$).}
\end{minipage}

\vspace{1.2em}

% ----------------------------------------------------------------------
% Task 30: Cubic f(n) = n^3 + n^2 + n
% ----------------------------------------------------------------------
\subsubsection{Cubic task: \texorpdfstring{$f(n)=n^3+n^2+n$}{f(n) = n\textasciicircum 3+n\textasciicircum 2+n} (task 30)}\label{subsubsec:task30}

\noindent\begin{minipage}[t]{0.48\textwidth}
\textbf{Emergent self-replication} \par\vspace{0.3em}
\textbf{Hexadecimal tape (32 bytes):} \par
{\scriptsize\ttfamily\raggedright
\textcolor{zrep}{20 9A 5E 14 4B ED B0} \textcolor{zinit}{8A BD 67 D3 9B 42 20 85} \textcolor{zunused}{DF} \par
\textcolor{zunused}{C6 37 7E E8} \textcolor{zarith}{94 10 3D} \textcolor{zhalt}{5F 20 E4 76} \textcolor{zunused}{DC 00 4D 3D C9} \par
} \vspace{0.4em}
\textbf{Disassembly:} \par
{\footnotesize\ttfamily\raggedright
\textcolor{zrep}{00: 20 9A    JR NZ, 1Ch   ; entry branch (not taken; ZF = 1 at entry)} \par
\textcolor{zrep}{02: 5E       LD E, (HL)   ; E = mem[00h] = 20h (partner offset)} \par
\textcolor{zrep}{03: 14       INC D        ; D = n + 1} \par
\textcolor{zrep}{04: 4B       LD C, E      ; BC = 32 (byte counter)} \par
\textcolor{zrep}{05: ED B0    LDIR         ; copy tape to partner half (20h--3Fh)} \par
\textcolor{zinit}{07: 8A       ADC A, D     ; A = n + 1 (FFh + (n+1) + CF)} \par
\textcolor{zinit}{08: BD       CP L         ; compare vs. L = 20h: sets NZ (A < 32)} \par
\textcolor{zinit}{09: 67       LD H, A      ; pass entry: H = current accumulator} \par
\textcolor{zinit}{0A: D3 9B    OUT (9Bh), A ; port write (no architectural effect)} \par
\textcolor{zinit}{0C: 42       LD B, D      ; B = n + 1 (loop counter)} \par
\textcolor{zinit}{0D: 20 85    JR NZ, 14h   ; jump to accumulation loop} \par
\textcolor{zunused}{0F--13:      DF C6 37 7E E8 [not executed on this path]} \par
\textcolor{zarith}{14: 94       SUB H        ; loop body: A = A - H} \par
\textcolor{zarith}{15: 10 3D    DJNZ 14h     ; loop n+1 times: A = A - (n+1)H = -n * A\_entry} \par
\textcolor{zhalt}{17: 5F       LD E, A      ; E = accumulator (pass 2: E = n\string^3 + n\string^2 + n)} \par
\textcolor{zhalt}{18: 20 E4    JR NZ, 3Eh   ; if A != 0: jump into replicated copy at 3Eh} \par
\textcolor{zhalt}{1A: 76       HALT         ; reached directly when A = 0, or via the dispatch below} \par
\textcolor{zunused}{1B--1F:      [Trailing bytes; their clone at 3Eh--3Fh decodes as DEC A; RET, whose stack pops re-enter 09h for the second pass (H updated to the new accumulator) and finally dispatch PC to the HALT at 1Ah]} \par
}
\end{minipage}
\hfill
\begin{minipage}[t]{0.48\textwidth}
\textbf{Hard-wired copy} \par\vspace{0.3em}
\textbf{Hexadecimal tape (32 bytes):} \par
{\scriptsize\ttfamily\raggedright
\textcolor{zinit}{A3 4B 04} \textcolor{zarith}{82 10 3D} \textcolor{zinit}{0C 43} \textcolor{zhalt}{5F} \textcolor{zinit}{9B 22 24 22 BC 08 D0} \par
\textcolor{zinit}{C2 97 0B B9 3A AE 98 BB 43 51 F8} \textcolor{zhalt}{76} \textcolor{zunused}{D6 61 19 13} \par
} \vspace{0.4em}
\textbf{Disassembly:} \par
{\footnotesize\ttfamily\raggedright
\textcolor{zinit}{00: A3       AND E        ; accumulator masking} \par
\textcolor{zinit}{01: 4B       LD C, E      ; seed register C} \par
\textcolor{zinit}{02: 04       INC B        ; inner loop counter} \par
\textcolor{zarith}{03: 82       ADD A, D     ; stage 1: accumulate n} \par
\textcolor{zarith}{04: 10 3D    DJNZ 03h     ; loop over quadratic term (target 43h == 03h mod 64)} \par
\textcolor{zinit}{06: 0C       INC C        ; outer counter adjust} \par
\textcolor{zinit}{07: 43       LD B, E      ; load partial product to B} \par
\textcolor{zhalt}{08: 5F       LD E, A      ; store intermediate in E} \par
\textcolor{zinit}{09: 9B       SBC A, E     ; residual adjustment} \par
\textcolor{zinit}{0A: 22 24 22 LD (2224h),HL; scratch write (target == 24h mod 64, partner half)} \par
\textcolor{zinit}{0D: BC       CP H         ; compare register H} \par
\textcolor{zinit}{0E: 08       EX AF, AF'   ; swap accumulator/flag set with shadow} \par
\textcolor{zinit}{0F: D0       RET NC       ; stack dispatch on carry-clear} \par
\textcolor{zinit}{10: C2 97 0B JP NZ, 0B97h ; conditional dispatch (target == 17h mod 64)} \par
\textcolor{zinit}{13: B9       CP C         ; compare with counter} \par
\textcolor{zinit}{14: 3A AE 98 LD A, (98AEh); A = mem[2Eh mod 64] (scratch half)} \par
\textcolor{zinit}{17: BB       CP E         ; status check} \par
\textcolor{zinit}{18: 43       LD B, E      ; set cubic multiplier} \par
\textcolor{zinit}{19: 51       LD D, C      ; store register} \par
\textcolor{zinit}{1A: F8       RET M        ; stack dispatch on sign-set} \par
\textcolor{zhalt}{1B: 76       HALT         ; halt execution} \par
\textcolor{zunused}{1C--1F:      [Unused trailing genomic bytes]} \par
}
\end{minipage}

\vspace{0.4em}
\noindent\begin{minipage}[t]{0.48\textwidth}
\textit{\small Validation cost: $\sim$88 CPU steps (mean over $x \in \{0, \dots, 15\}$).}
\end{minipage}\hfill
\begin{minipage}[t]{0.48\textwidth}
\textit{\small Validation cost: 44 CPU steps (mean over $x \in \{0, \dots, 15\}$).}
\end{minipage}

\vspace{1.5em}

% ======================================================================
% Section 3: Emergence of Horner-style evaluation in 64-task environments
% ======================================================================
\section{Emergence of Horner-style polynomial evaluation in high-degree environments}\label{app:horner_evaluation}

To test whether the 32-byte limit prevents the evolution of higher-degree polynomial solvers beyond the cubic programs analyzed in Sec.~\ref{app:discovered_architectures} (and Secs.~\ref{subsubsec:task28}, \ref{subsubsec:task30}), we ran an extended experiment expanding the environment to 64 task niches (building on the niche framework described in Secs.~\ref{subsec:model_overview}, \ref{subsec:cnp}, and \ref{subsec:methods_competence}), introducing quartic, quintic, and sextic polynomial targets. Under this expanded regime, the population evolved an exact solver for the sextic target $f(n)=n^6+n^2+n$.

\subsection{Algorithmic structure and Horner recurrence}\label{subsec:horner_recurrence}

Evaluating higher-degree polynomials within 32 bytes presents a space constraint. Evaluating individual terms $a_k n^k$ separately would require computing and storing intermediate powers of $n$ ($n^2$, $n^3$, $n^4$, $n^5$), which requires more registers or scratchpad memory than is available in a 32-byte genome.

Instead, the evolved program implements the nested factorization of Horner's method~\citep{Horner1819}. For $f(n)=n^6+n^2+n$, Horner's nested form factors the polynomial as:
\begin{equation}
   f(n) = n \cdot \big( 1 + n \cdot ( 1 + n^4 ) \big) = n^6 + n^2 + n.
\end{equation}
This recurrence evaluates the polynomial in five successive outer multiplication passes, updating a single running accumulator without storing intermediate monomial powers. Tracing the register states across iterations gives intermediate values that match the Horner partial sums for the coefficient vector $(a_0, \dots, a_6) = (0, 1, 1, 0, 0, 0, 1)$:
\begin{equation}
\begin{aligned}
   s_0 = 1 &\;\longrightarrow\; s_1 = n \;\longrightarrow\; s_2 = n^2 \;\longrightarrow\; s_3 = n^3 \\
           &\;\longrightarrow\; s_4 = n^4 + 1 \;\longrightarrow\; s_5 = n^5 + n + 1 \;\longrightarrow\; s_6 = n^6 + n^2 + n.
\end{aligned}
\end{equation}
The first partial sum $s_1 = n$ is obtained by seeding the accumulator directly with the input, so the five multiplication passes take $s_1$ through $s_6$.

\subsection{Z80 execution mechanics and return-oriented dispatch}\label{subsec:horner_mechanics}

The evolved program implements this recurrence through several interconnected mechanisms (detailed in Table~\ref{tab:horner_sextic_listing}):
\begin{itemize}
   \item \textbf{Outer loop control via signed overflow:} The number of outer iterations is governed by repeatedly subtracting the constant \texttt{0x2D} (45 decimal) from accumulator \texttt{A} (\texttt{SBC A, 2Dh}), initialized to \texttt{0xFF}. In the Z80 architecture, 8-bit arithmetic subtractions update the parity/overflow ($P/V$) flag to reflect signed two's complement overflow ($V$). A conditional return instruction (\texttt{RET PO}) acts as a state dispatcher, executing returns while $V=0$ and falling through when signed overflow ($V=1$) occurs.
   \item \textbf{Inner multiplication kernel:} Within each outer iteration, an inner loop comprising \texttt{ADD HL, DE} (\texttt{0x19}) and \texttt{DJNZ} (\texttt{0x10}) computes multiplication by repeated addition, accumulating $HL \leftarrow HL + n \cdot DE$ in $n$ steps.
   \item \textbf{Coefficient injection via \texttt{INC L}:} On two of the five passes (the third and fourth, as observed in emulation), execution falls through \texttt{RET PO} and reaches an \texttt{INC L} instruction (\texttt{0x2C}), injecting a $+1$ seed into the low byte of the accumulator. Subsequent multiplication passes scale these seeds into the lower-order terms $n^2 + n$. Which passes fall through is determined jointly by the signed-overflow trajectory of the \texttt{SBC} updates and by the stack shift introduced by \texttt{DEC SP} (below), and is therefore established by execution rather than by the flag arithmetic alone.
   \item \textbf{Return-oriented dispatch:} As with the shorter programs in Sec.~\ref{app:discovered_architectures}, during self-replication (\texttt{LDIR}), the 32-byte tape is cloned into the upper half of memory (\texttt{0x20}--\texttt{0x3F}). When \texttt{RET PO} pops a return address from the stack (initialized as in Sec.~\ref{subsec:methods_competence}, resolving to the last byte of the 64-byte space), it reads 16-bit word addresses directly from the cloned genome. The genome thus serves simultaneously as executable instructions and as a static return-address dispatch table. Because every fall-through also executes \texttt{DEC SP}, successive pops read from shifted genome offsets, so the effective dispatch targets change over the course of a run.
\end{itemize}

\subsection{Execution complexity}\label{subsec:horner_complexity}

For inputs $n \ge 1$, the total instruction execution count of the evolved sextic solver follows the linear relation:
\begin{equation}
   C(n) = 88 + 10n.
\end{equation}
Each of the five outer multiplication iterations contains the inner \texttt{ADD}/\texttt{DJNZ} pair executed $n$ times, contributing $5 \times 2 \times n = 10n$ instruction cycles. The constant term 88 accounts for the initial 32-byte self-replication preamble (\texttt{LDIR}), register configuration, and outer loop dispatch bookkeeping. Linearity in the input $n$ confirms that the program executes exactly five linear multiplication stages rather than generating independent higher-degree monomial subroutines.

\subsection{Degree parameterization via constant substitution}\label{subsec:horner_parameterization}

The degree of the evaluated polynomial is controlled by the operand of the \texttt{SBC A, 2Dh} instruction at offset \texttt{0x18}. Together with the initialization $A = \texttt{0xFF}$, this constant fixes the trajectory of the signed-overflow flag across successive state updates, and thereby determines both the number of multiplication passes executed before \texttt{HALT} and the subset of passes on which execution falls through \texttt{RET PO} into the \texttt{INC L} coefficient injection. As noted in Sec.~\ref{subsec:horner_mechanics}, this trajectory is not a closed-form function of the repeated subtraction alone, because each fall-through executes \texttt{DEC SP} and shifts all subsequent return-address dispatch; the effective injection schedule is therefore determined by execution.

Substituting the single control byte \texttt{0x8D} (141 decimal) changes the state update to $A \leftarrow A - 141 - \mathrm{CF}$. In emulation, the modified program performs 13 multiplication passes before reaching \texttt{HALT} and evaluates
\begin{equation}
   P(n) = n^{14} + n^{11} + n^9 + n^7 + n^6 + n^5 + n^4 + n^2 + 1 \pmod{256},
\end{equation}
with intermediate register values across all 13 passes matching the Horner partial sums of this 14th-degree polynomial. A single-byte change in the evolved genome thus retunes the program from a sextic solver into a 14th-degree polynomial evaluator.

\begin{table}[htbp]
\centering
\small
\caption{\textbf{Annotated disassembly of the evolved 32-byte sextic polynomial solver ($n^6+n^2+n$).} The 32-byte program implements a Horner-style nested multiplication accumulator governed by signed overflow transitions ($V$) and return-oriented stack dispatch. All jump targets, memory operands, and stack accesses resolve modulo the 64-byte address space.}
\label{tab:horner_sextic_listing}
\begin{tabular}{@{}lllp{7.2cm}@{}}
\toprule
\textbf{Offset} & \textbf{Hex bytes} & \textbf{Instruction} & \textbf{Functional role} \\
\midrule
\texttt{00--01} & \texttt{20 2B}    & \texttt{JR NZ, 2Dh}       & Entry branch (falls through when $ZF=1$) \\
\texttt{02}     & \texttt{5E}       & \texttt{LD E, (HL)}       & Load partner destination address ($E \leftarrow$ \texttt{mem[00h]} $=$ \texttt{20h} $= 32$) \\
\texttt{03}     & \texttt{4B}       & \texttt{LD C, E}          & Set byte counter ($BC \leftarrow 32$) \\
\texttt{04--05} & \texttt{ED B0}    & \texttt{LDIR}             & Self-replication: copy 32-byte tape to partner cell \\
\texttt{06}     & \texttt{BD}       & \texttt{CP L}             & Clear carry flag; initialize flags for loop entry \\
\texttt{07--09} & \texttt{C3 14 4C} & \texttt{JP 4C14h}         & Jump to loop entry pipeline (target $\equiv$ \texttt{14h} mod 64) \\
\texttt{0A}     & \texttt{37}       & \texttt{SCF}              & Spacer byte (not executed on observed paths) \\
\texttt{0B--0C} & \texttt{D3 5A}    & \texttt{OUT (5Ah), A}     & Dual-use: port write as an instruction; its operand byte \texttt{5Ah} (offset \texttt{0Ch}) decodes as \texttt{LD E, D} when reached via dispatch, re-seeding the accumulator \\
\texttt{0D}     & \texttt{44}       & \texttt{LD B, H}          & Load inner multiplication loop counter ($B \leftarrow n$) \\
\texttt{0E--10} & \texttt{EA BA 76} & \texttt{JP PE, 76BAh}     & Conditional dispatch; its operand byte \texttt{76h} (offset \texttt{10h}) doubles as the program's \texttt{HALT} when reached as an instruction via dispatch \\
\texttt{11}     & \texttt{19}       & \texttt{ADD HL, DE}       & Inner multiplication stage: $HL \leftarrow HL + DE$ \\
\texttt{12--13} & \texttt{10 FD}    & \texttt{DJNZ 11h}         & Loop inner addition $n$ times ($HL \leftarrow HL + n \cdot DE$) \\
\texttt{14}     & \texttt{EB}       & \texttt{EX DE, HL}        & Pipeline swap: transfer accumulated product to $DE$ \\
\texttt{15}     & \texttt{69}       & \texttt{LD L, C}          & Reset low accumulator byte ($L \leftarrow 0$; $C = 0$ after \texttt{LDIR}) \\
\texttt{16}     & \texttt{54}       & \texttt{LD D, H}          & Set multiplicand ($D \leftarrow n$) \\
\texttt{17--18} & \texttt{DE 2D}    & \texttt{SBC A, 2Dh}       & State update: $A \leftarrow A - 45 - \mathrm{CF}$; sets signed overflow ($V$) \\
\texttt{19}     & \texttt{E0}       & \texttt{RET PO}           & Stack dispatch: pop return target when $V=0$ \\
\texttt{1A--1C} & \texttt{CC 52 70} & \texttt{CALL Z, 7052h}    & Unreached branch on fallthrough ($ZF=0$) \\
\texttt{1D}     & \texttt{B5}       & \texttt{OR L}             & Status normalization on overflow branch \\
\texttt{1E}     & \texttt{3B}       & \texttt{DEC SP}           & Stack shift: re-aims subsequent dispatch pops \\
\texttt{1F}     & \texttt{2C}       & \texttt{INC L}            & Coefficient injection: inject $+1$ seed into accumulator \\
\bottomrule
\end{tabular}
\end{table}

\end{document}